%% file: trapping.tex
\documentclass[11pt,draftcls,onecolumn]{IEEEtran}

\usepackage{multicol}
\usepackage{subfigure}
\usepackage{times}
\usepackage[cmex10]{amsmath}
\usepackage{epsfig,cite,alltt,epic,array,multicol,graphicx}
\usepackage{wasysym}
\usepackage{pstricks}
\usepackage{multirow}
\usepackage{subfigure}
\usepackage{array}
\graphicspath{{./}{figs/}}
\usepackage{threeparttable}
\usepackage{footnote}
\makesavenoteenv{tabular}

\begin{document}
\title{Three Efficient, Low-Complexity Algorithms for Automatic Color Trapping\thanks{This research was supported by the Hewlett-Packard Company.}}
\vspace{-0.2in}
\author{Haiyin~Wang,~\IEEEmembership{Student Member,~IEEE,}
        Mireille~Boutin,~\IEEEmembership{Member,~IEEE,}
        Jeffrey~Trask,~\IEEEmembership{}and~Jan~Allebach,~\IEEEmembership{Fellow,~IEEE}
\thanks{H.~Wang, M.~Boutin, and J.~Allebach are with the School
of Electrical and Computer Engineering, Purdue University, West Lafayette,
IN, 47907 USA e-mail: hw, mboutin, allebach @purdue.edu.}
\thanks{J.~Trask is with the Hewlett-Packard Company, 11311 Chinden Blvd. Boise, ID 83714.}
\thanks{This research was described in part in H. Wang, M. Boutin, J. Trask and J. Allebach, ``An efficient method for color trapping", in {\it Color Imaging XIII: Processing, Hardcopy, and Applications,  IS\&T/SPIE Electronic Imaging Symposium}, San Jose, CA, Jan-Feb 2008. }
}
\maketitle

\vspace{-0.4in}
\begin{abstract}
Color separations (most often cyan, magenta, yellow, and black) are
commonly used in printing to reproduce multi-color images. For
mechanical reasons, these color separations are generally not
perfectly aligned with respect to each other when they are rendered
by their respective imaging stations. This phenomenon, called color
plane {\em misregistration}, causes gap and halo artifacts in the
printed image. Color trapping is an image processing technique that
aims to reduce these artifacts by modifying the susceptible edge
boundaries to create small, unnoticeable overlaps between the color
planes (either at the page description language level or the
rasterized image level). In this paper, we propose three
low-complexity algorithms for automatic color trapping at the
rasterized image level which hide the effects of small color plane
mis-registrations. Our proposed algorithms are designed for software
or embedded firmware implementation. The trapping method they follow
is based on a hardware-friendly technique proposed by J.~Trask
(JTHBCT03) which is too computationally expensive for software or
firmware implementation. The first two algorithms are based on the
use of look-up tables (LUTs). The first LUT-based algorithm corrects
all registration errors of one pixel in extent and reduces several
cases of misregistration errors of two pixels in extent using only
$727$ Kbytes of storage space. This algorithm is particularly
attractive for implementation in the embedded firmware of low-cost
formatter-based printers. The second LUT-based algorithm corrects
all types of misregistration errors of up to two pixels in extent
using $3.7$ Mbytes of storage space. This algorithm is more suitable
for software implementation on host-based printers. The third
algorithm is a hybrid one that combines look-up tables and feature
extraction to minimize the storage requirements ($724$ Kbytes) while
still correcting all misregistration errors of up to two pixels in
extent. This algorithm is suitable for both embedded firmware
implementation on low-cost formatter-based printers and software
implementation on host-based printers. All three of our proposed
algorithms run, in average, more than three times faster than a
software implementation of JTHBCT03.
\end{abstract}

\begin{IEEEkeywords}
Color trapping, color plane misregistration, look-up tables.
\end{IEEEkeywords}

\input{introduction}
\input{background}
\input{trask_approach}
\input{LUT-based}
\input{hybrid}
\input{conclusion}


\bibliographystyle{IEEEtran}
\bibliography{IEEEabrv,ColorTrappingRef}





\end{document}

%% file: introduction.tex
\section{Introduction}
\label{Intro}

\begin{figure}[tbp]
\centerline{\epsfig{figure=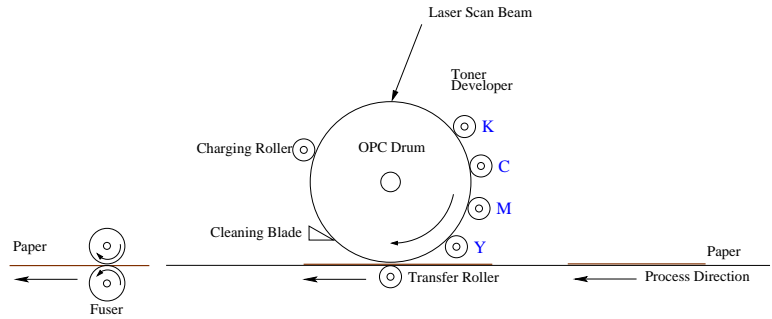, width = 4in}}
\caption{Multi-pass color EP printer architecture. The different
colors of toner are sequentially applied to the OPC drum surface
before being transferred in a single step to the paper and
subsequently fused.} \label{fig:multipass}
\end{figure}

\IEEEPARstart{T}{he} color laser printer market is currently
dominated by two laser electrophotographic (EP) printing
architectures: multi-pass and in-line. Multi-pass color printers
(Fig. \ref{fig:multipass}) operate by sequentially overlaying single
color image planes on an optical photoconductor (OPC) drum and
subsequently transferring all image planes in a single step onto the
paper. The surface of the OPC drum thus acts as an intermediate
transfer material on which all the different colorants (i.e., toner
powders) are applied. This is done by first charging the OPC drum
surface using a charging roller and putting a toner unit into
position. A laser beam is then used to selectively remove the charge
on the OPC drum surface according to the image for that color plane.
The locations where the charges have been removed attract the toner
particles to the OPC drum surface. This process is repeated four
times - one time each for cyan ($C$), magenta ($M$), yellow ($Y$),
and black ($K$). The OPC drum is subsequently put in contact with
the paper to transfer the colored particles onto it, and a cleaning
blade is applied to the OPC drum. The paper then goes through a
fuser, which is used to bond all four colors of toner to the paper
by exerting heat and pressure. Since the image planes are created in
separate mechanical operations within the printing process, it is
very difficult and costly to perfectly align them.

\begin{figure}[tbp]
\centerline{\epsfig{figure=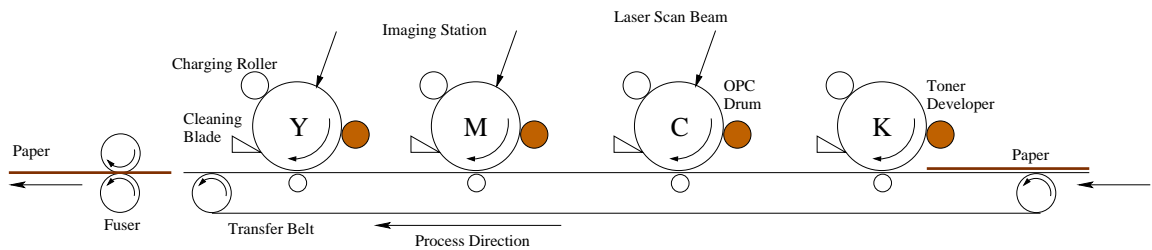, width = 6in}}
\caption{In-line color EP printer architecture. Separate imaging
stations are used to apply a single colorant to the paper (or an
intermediate transfer material).} \label{fig:inline}
\end{figure}

For in-line printers (Fig. \ref{fig:inline}), there are four
separate imaging stations, each with its own OPC drum and toner unit
($C$, $M$, $Y$, or $K$). The toner is either applied directly on the
paper itself while it is transported by a transfer belt, or the
transfer belt is used as an intermediate transfer material. All four
toner colors are fused onto the paper in a single final step.
Misregistration occurs as a result of misalignments between the
paper (or intermediate transfer material) and the imaging stations.

\begin{figure}[!t]
\centerline{\epsfig{figure=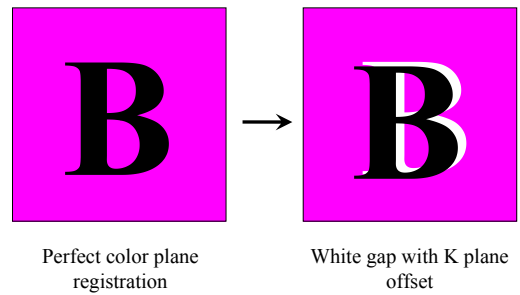, height = 1.5in}}
\caption{Example of white gap artifact. Black text ($100\%$ $K$) is
printed on a magenta ($100\%$ $M$) background. If the $K$ plane is
slightly shifted, a white gap appears around the text.}
\label{fig:white gap}
\end{figure}

\begin{figure}[!t]
\centerline{\epsfig{figure=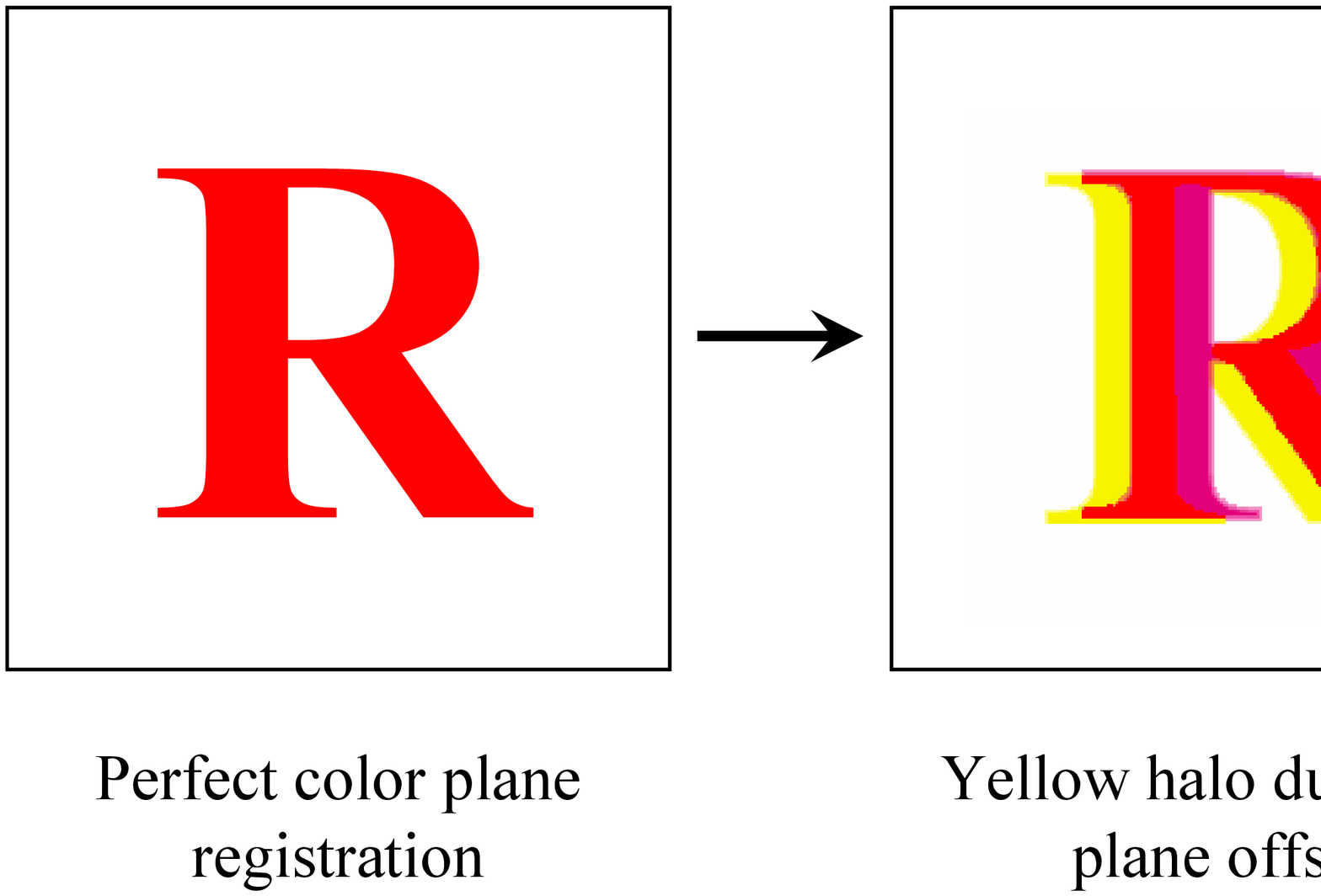, height = 1.5in}}
\caption{Example of yellow halo artifact. Red text ($100\%$ $M$,
$100\%$ $Y$) is printed on a white background. If the $M$ plane is
slightly shifted, a yellow halo appears around the
text.}\label{fig:yellow halo}
\end{figure}

When printed at $600$ dpi resolution, misalignments of a magnitude
as small as one pixel can form visible artifacts in the printed
image. A mechanical accuracy of more than $1/1000$ inch in the image
plane position is required to prevent visible artifacts~\cite{Tr03}.
Unfortunately, desktop, workgroup, and office printers do not
possess this level of mechanical accuracy. This results in gap
and/or halo artifacts near the edges of objects in each color plane.
Examples of a white gap artifact and a yellow halo artifact are
shown in Figs. \ref{fig:white gap} and \ref{fig:yellow halo},
respectively.

Color trapping is a workaround to this problem that consists of
moving the edge boundaries of the lighter colorants underneath the
edge boundaries of the darker colorants. While the resulting change
in the image is almost imperceptible to the human eye, this prevents
the appearance of gap and halo artifacts caused by small color plane
mis-registrations. Color trapping is commonly used in high-quality
commercial printing with an offset press. Typically, this is done
manually by a trained graphic artist using a professional page
layout application. Manual color trapping entails examining each
page displayed on a computer screen to predict where the
registration errors are likely to occur, and creating {\em traps} to
prevent the errors. Until recently, the underlying professional
printing applications were quite expensive. More recently, a class
of low-cost desktop publishing software (e.g., Adobe InDesign, Quark
XPress) has emerged that allows users with a personal computer to
create high quality documents. Still, this manual procedure is
tedious and requires advanced skills, which makes it unattractive
for low-end publishing by relatively casual users. A fully automatic
solution is thus preferred.

There is a very limited body of literature on automatic color
trapping. As far as we know, this issue has not previously been
discussed in the scholarly engineering literature. Aside from
nondescriptive commercial product advertisements, the only published
work on the subject appears to be in the form of patents. We give a
high-level summary of current automatic color trapping approaches in
the next section.

Automatic color trapping can be performed in hardware, in software,
or in firmware. Hardware-based color trapping is done by including
specific hardware components (e.g., printer application specific
integrated circuits (ASICs)) to perform specific tasks within the
printer. Once a hardware circuit has been designed, it is difficult
to alter. So while hardware-based trapping is fast, it is inflexible
and costly. Software-based trapping is performed by the host
computer, either through a stand-alone application such as Adobe
InDesign, or through the printer driver. Software-based trapping is
flexible, easy to tailor for different trapping needs, and also cost
effective. Software friendly algorithms that have low memory and
computational requirements can also be implemented directly in
firmware to run on the microprocessor of a formatter-based printer.

Due to architectural differences, hardware-based algorithms are
typically not suitable for software implementation and vice versa.
In this paper, we present three efficient automatic color trapping
algorithms for software or firmware implementation. Our approaches
build on a hardware-friendly algorithm developed by
J.~Trask~\cite{Tr03} in 2003 (JTHBCT03). Overall, the trapped images
they produce are very similar to those generated by JTHBCT03.
However, all three proposed algorithms are computationally simpler
than JTHBCT03. They are thus more amenable to software
implementation. In addition, two of them have very low memory
requirement, which makes them suitable for firmware implementation.

JTHBCT03 uses a $5\times5$ sliding window to determine how to
process each pixel of a given rasterized image. One approach that we
use to reduce the complexity is to employ look-up tables (LUTs). In
particular, in all three algorithms, the trapping parameters, which
determine how the color of the center pixel of the window is to be
modified, are stored in a LUT rather than being computed every time.
In our first algorithm, an approximation of JTHBCT03 is obtained by
considering a smaller ($3\times3$ pixel) window. The pixel
configuration is used to obtain the index of a LUT in which is
stored the corresponding action that should be taken: either trap as
an edge pixel or do not trap. This algorithm reduces several cases
of misregistration errors up to two pixels in extent using only
$727$ Kbytes of storage space. It is particularly attractive for
implementation on the firmware of low-cost formatter-based printers.

In our second algorithm, a $5\times5$ window is considered and a
series of LUTs are built to replicate JTHBCT03's decision: either
trap as an edge pixel, trap as a neighboring edge pixel, or do not
trap. This algorithm corrects all types of misregistration errors up
to two pixels in extent using $3.7$ Mbytes of storage space. While
the memory requirement is too high for firmware implementation on
low-cost formatter-based printers, this algorithm is suitable for
software implementation on host-based printers.

In our third algorithm, the rather large LUTs of the second
algorithm are replaced with a hybrid approach using feature
extraction together with some small LUTs. For example, some simple
features are used to identify the majority of non-trappable pixels
before they even enter the classification stage. The rules that
determine whether and how a pixel should be trapped are also stored
in a small LUT based on three simple discrete features. The decision
boundaries corresponding to the latter LUT can be easily visualized
and modified for different trapping requirements. Overall, this
approach allows us to significantly decrease the number of ``if''
statements, additions, and multiplications as well as the overall
CPU time required to trap a typical page without requiring a large
storage capacity. This algorithm is suitable for both firmware
implementation on low-cost formatter-based printers and software
implementation on host-based printers. All three of our proposed
algorithms run more than three times faster than a software
implementation of JTHBCT03.

The remainder of this paper is organized as follows.
Sec.~\ref{background} gives a high-level summary of the existing
automatic color trapping methodologies. Section~\ref{Trask's
approach} presents the details of JTHBCT03. Our two LUT-based color
trapping algorithms (Algorithm 1 and Algorithm 2) are explained in
Sec.~\ref{LUT approaches}, where we begin by describing the
straightforward {\em pixel-independent} approach, before introducing
the computationally simpler {\em pixel-dependent} approach. In
Sec.~\ref{hybrid method} we present our hybrid algorithm (Algorithm
$3$), which uses feature extraction along with LUTs. We summarize
our results and conclude in Sec.~\ref{conclusion}.

%% file: background.tex
\section{Color Trapping Overview}
\label{background}

Color trapping is a process in which color edges are either expanded
or shrunk to create an overlap of colors to prevent small
registration errors from causing gap or halo artifacts. Most
automatic color trapping methods (e.g.,
\cite{Kl01,WeFu01,Yh00,Mo99,DeBjBlBePeRo96,Bl94,DeRe93,De93}) are
{\em object-based}, in the sense that they analyze the
representation of a printed page (e.g., page-description-language
(PDL) or structured graphics) in order to obtain information about
the objects, detect where the edges are, and then perform the
trapping accordingly. The trapping is done independently of the
output resolution of the printing device. In contrast, JTHBCT03 is a
{\em pixel-based} method, which modifies the image at the printing
stage, after the output resolution has been determined and the page
has been rasterized.

Object-based trapping has several advantages. First, it is
independent of the originating program: as long as the printed page
is expressed in PDL format, it can be trapped. Second, only the edge
pixels are considered instead of the entire set of pixels in the
frame buffer. Third, the number of edge pixels does not grow as the
resolution increases; and it is independent of the number of
separation colorants. However, the interaction between edges and
objects can be quite complex, especially when there are many color
objects involved. Moreover, the operations involved in the vector
processing itself also tend to be complicated. This complexity
increases dramatically as the number of edges increases.

We chose the pixel-based color trapping approach because of its
simplicity, as it is applied directly on a raster (bit-map) image
generated at the desired output resolution. The raster image is
trapped in a local fashion based on the actual pixel data, which
yields a more straightforward algorithm. This requires each color
separation or a swath thereof to be stored as an individual plane in
a frame buffer. The planes of the frame buffer are then trapped
pixel by pixel, and the combined results determine the trapped
image. Determining the color transitions and edges is easier on a
rasterized image than from a high level page description because the
bit map explicitly describes the color of each pixel. Moreover, the
set of the operations needed to process a single pixel in a frame
buffer is always the same no matter what the printing resolution is.

%% file: trask_approach.tex
\section{Trask's 2003 Color Trapping Algorithm}
\label{Trask's approach}

\begin{figure}[!t]
\centering
\centerline{\epsfig{figure=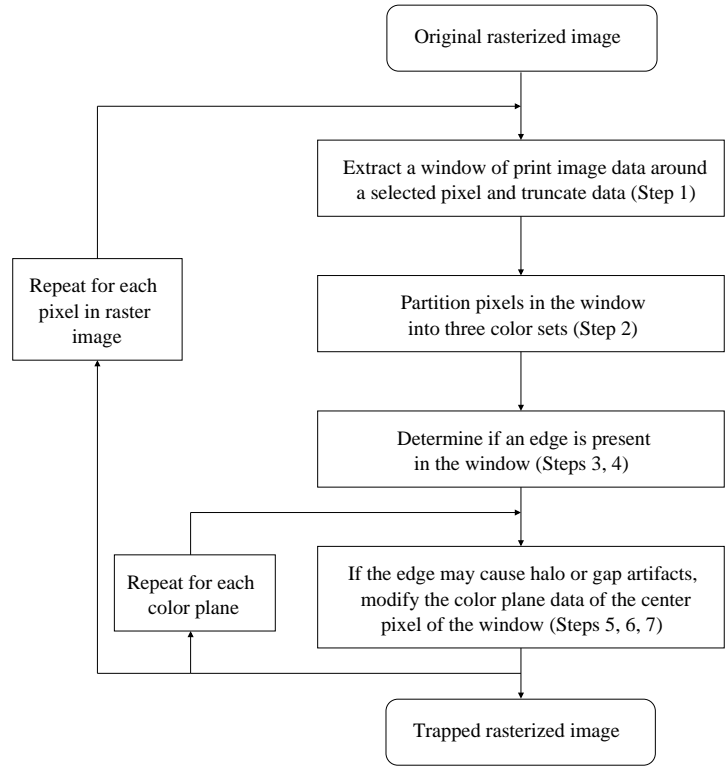,height=4.0in}}
\caption{The flow diagram of our software implementation of
JTHBCT03.} \label{fig:diagram}
\end{figure}

Our starting point is the pixel-based automatic color trapping
algorithm developed by J.~Trask \cite{Tr03} in 2003 (JTHBCT03). This
algorithm was especially designed for hardware implementation.
Unfortunately, when implemented in software, it is too time
consuming for most applications. For example, processing a single
$600$ dpi page typically takes longer than $24$ seconds on a
computer with an Intel\textregistered \textrm{ }Xeon(TM) processor
and CPU speed of $3.60$ GHz. Our goal is to develop an algorithm
which accomplishes trapping in a way that is comparable to JTHBCT03,
but runs significantly faster in software. JTHBCT03 comprises seven
steps, which we now summarize. An illustration is provided in
Fig.~\ref{fig:diagram}.

\subsection*{Step One: Bit Truncation}
\label{color discretization} In this step, a $5\times5$ window of
the $C$, $M$, and $K$ color values are extracted and truncated into
$5$ bits each. This reduces the storage requirement when the
algorithm is implemented in ASICs and also makes the algorithm more
robust to small color value changes due to noise.

\subsection*{Step Two: Color Categorization}
\label{color categorization} 
\begin{figure}[!t]
\centerline{ \scalebox{0.8}{
\input{./figs/tolerance_vol.pstex_t}
} } \caption{Tolerance volume for a color $A$ pixel. It is computed
based on the $C$, $M$, and $K$ color values of the center pixel
(represented by the red dot) of the $5\times5$ window.}
\label{fig:tolerance_vol}
\end{figure}
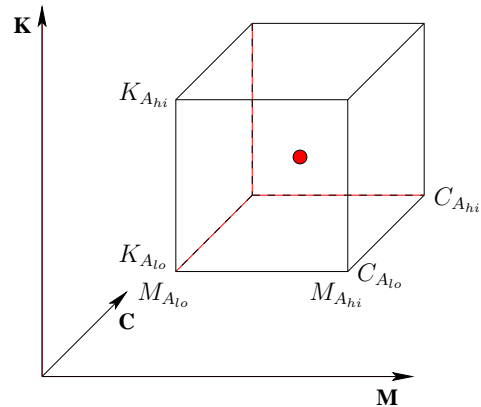

\begin{figure}[!t]
\centerline{\epsfig{figure=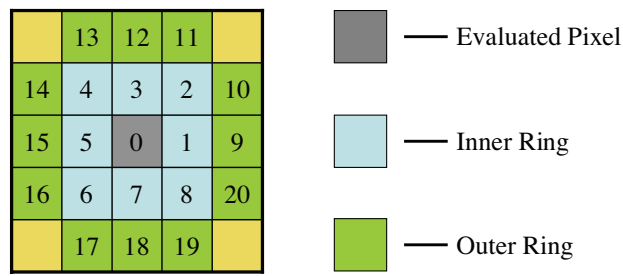,height=1.4in}}
\caption{Pixel scan order for a $5\times5$ window. The pixel
numbered $0$ represents the pixel to be evaluated by the trapping
process. The pixels numbered $1-8$ represent the inner ring; and the
pixels numbered $9-20$ represent the outer ring of the $5\times5$
window. The four corner pixels are ignored by the trapping algorithm
since their distance to the center pixel is greater than two pixels.}
\label{fig:scan_order}
\end{figure}

The center pixel of the window (now
represented by truncated data) is labeled as color $A$. Then the
{\em tolerance volume} is computed. This volume encloses all colors
which look similar enough to the center pixel color to be
categorized as $A$ pixels as well. The dimensions of the tolerance
volume are illustrated in Fig. \ref{fig:tolerance_vol}. All the
pixels in the $5\times5$ window are scanned according to the order
described in Fig.~\ref{fig:scan_order}, and the first pixel falling
outside of the tolerance volume is categorized as a color $B$ pixel.
The tolerance volume for the color $B$ pixel is then computed. All
pixels falling outside of both tolerance volumes for color $A$ and
color $B$ pixels are categorized as color $O$ (other) pixels.

\subsection*{Step Three: Feature Extraction}
\label{feature extraction} The number and arrangement of color $A$,
$B$, and $O$ pixels in the window are characterized by $27$
features, which are extracted and stored. Among these features, $11$
describe the inner ring of the window (Fig. \ref{fig:scan_order}),
$12$ describe the outer ring, and $4$ describe the relationship
between the inner ring and the outer ring.

\subsection*{Step Four: Edge Detection}
\label{edge detection} When the window contains no edge, the center
pixel should not be trapped. When the window does contain an edge,
the center pixel should be trapped differently depending on whether
it lies on the edge or slightly away from the edge, and depending on
the colors of the two regions divided by the edge. So the center
pixel of the window is classified into one of four categories: {\em
edge1}, {\em edgey}, {\em edge2}, and {\em non-trappable}, which we
now describe briefly.

\subsubsection{$edge1$} \label{edge1} 
\begin{figure}[!t]
\centerline{\epsfig{figure=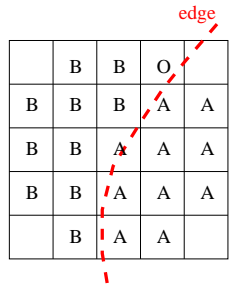, height = 1.5 in}}
\caption{An example of an {\em edge1} type pixel. The center pixel
of the $5\times5$ window resides on an edge between color $A$ and
color $B$ pixels.}\label{fig:edge1}
\end{figure}

The center pixel is directly on an edge
between color $A$ and color $B$ pixels. This type of edge may cause
gap or halo effects if the color planes are misaligned. An example
is shown in Fig. \ref{fig:edge1}.

\subsubsection{$edge2$}
\label{edge2} 
\begin{figure}[!t]
\centerline{\epsfig{figure=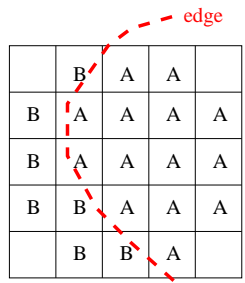, height = 1.5in}}
\caption{An example of an {\em edge2} type pixel. The center pixel
of the $5\times5$ window is one pixel away from an edge between
color $A$ and color $B$ pixels.}\label{fig:edge2}
\end{figure}

The center pixel of the sliding window is one pixel
away from an edge. This type of edge may also cause gap or halo
artifacts if the color planes are mis-aligned by two pixels. An
example is shown in Fig.~\ref{fig:edge2}.

\subsubsection{$edgey$}
\label{edgey} The center pixel of the sliding window is right on an
edge between two colors more saturated than those of {\em edge1}
type. So the pixel configurations are also similar to those of {\em
edge1} type, but with more color $A$ and $B$ pixels. For example, in
Fig~\ref{fig:edge1}, the color $O$ pixel is replaced with a color
$B$ pixel to represent an {\em edgey} type pixel. This type of edge
is detected when a color region of saturated red or green is
adjacent to a white region, which may lead to yellow halo artifacts
(Fig. \ref{fig:yellow halo}) if the color planes are misaligned. The
halo artifacts can be effectively reduced or eliminated if the $Y$
area is shrunk inside the area of the darker colorant. This
effectively pulls the white background color under this darker
colorant. Since we found that this type of edge is trapped in a
manner that is similar to the way in which {\em edge1} type is
trapped, we chose for simplicity to merge this category with the
{\em edge1} category.


The categorization described above is based on a series of condition
checks which uses the $27$ features extracted in Step Three. As many
as $76$ conditions may have to be checked in order to classify a
given pixel. When implemented in software, this step, along with the
previous one, are computationally expensive. In the next sections,
we propose alternative ways to make this decision. In particular,
the method proposed in Sec.~\ref{hybrid method} uses only three
features together with a small LUT.

\subsection*{Step Five: Color Density Calculation}
\label{density cal} Once a trappable edge is detected, the
approximate relative darkness of the applied colorant is calculated
within the context of the printed image. Usually a black color will
appear darker to the human eye than a cyan, magenta, or yellow with
the same level. So a weighted sum of the $C$, $M$, $Y$, and $K$
values of the evaluated pixel is used to represent the density
values. The details of the density calculation can be found
in~\cite{Tr03}.

\subsection*{Step Six: Trapping Parameter Calculation}
\label{trapping parameter cal} 
\begin{figure}[tbp]
\centerline{\epsfig{figure=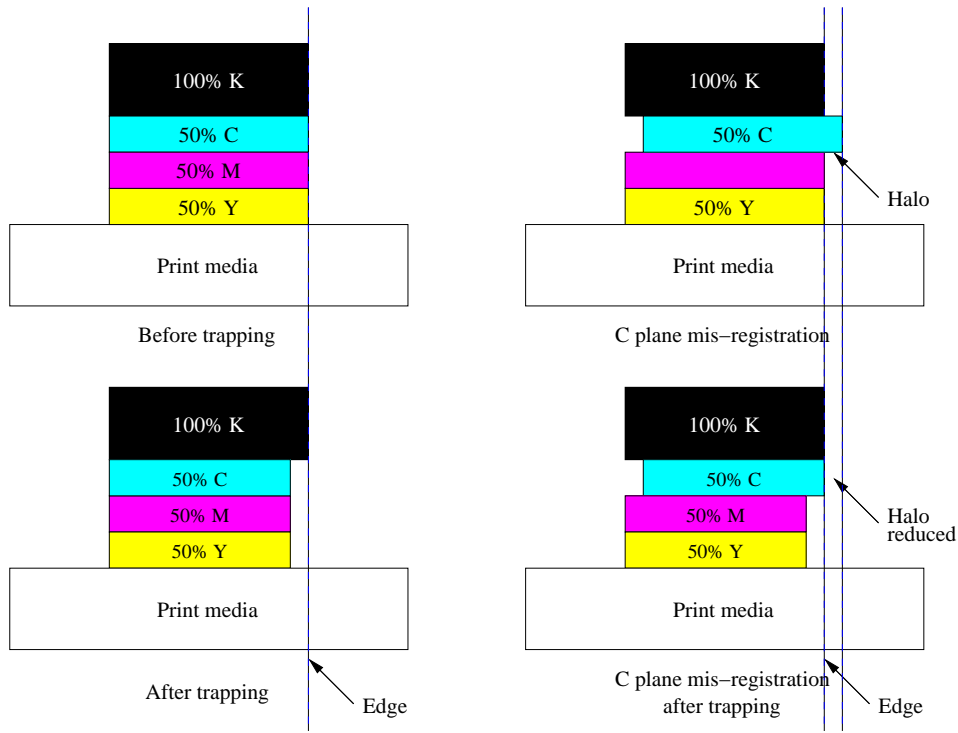, width = 5in}}
\caption{An example of halo reduction for $C$ plane misregistration.
Before trapping, a halo artifact will occur if the $C$ plane is
shifted to the right. To avoid the misregistration error, the $50\%$
$C$, $50\%$ $M$, and $50\%$ $Y$ planes are shrunk under the $K$
plane to eliminate the halo.}\label{fig:halo reduction}
\end{figure}

\begin{figure}[!t]
\centerline{\epsfig{figure=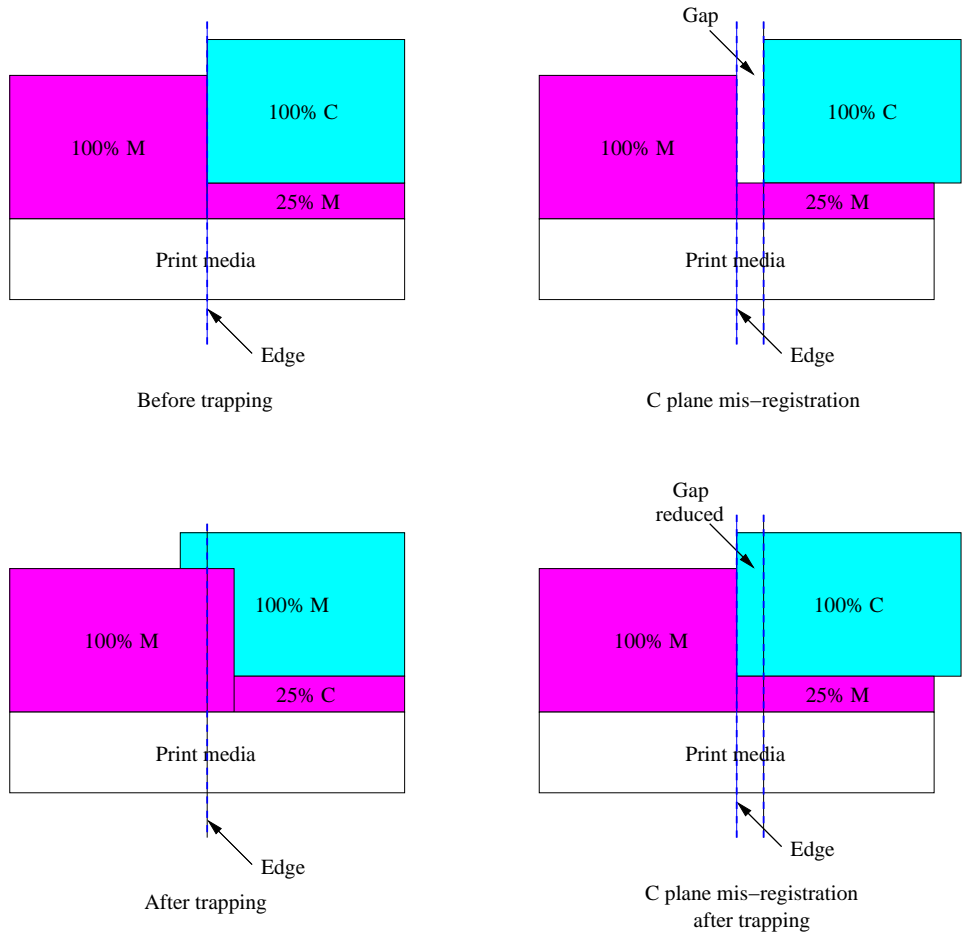, width = 5in}}
\caption{An example of gap reduction for $C$ plane misregistration.
Before trapping, a gap artifact will occur if the $C$ plane is
shifted to the right. To avoid the misregistration error, the
$100\%$ $C$ and $100\%$ $M$ planes are expanded to cover the
potential gap area.}\label{fig:gap reduction}
\end{figure}

The color densities that were
calculated in the previous step are then used to determine the
amount and type of trapping to be applied. As a general rule, in
order not to change the outline of the color object, the darker
color that forms the contour should not be changed after trapping.
Therefore, the lighter color regions are usually extended into the
darker ones to reduce the potential gap or halo artifacts.
Figs.~\ref{fig:halo reduction} and \ref{fig:gap reduction}
illustrate how gap and halo effects are reduced or eliminated.

\subsection*{Step Seven: Final Trapped Value Calculation}
\label{final trapped value cal} The final color plane values of the
center pixel in the window are calculated based on various selected
trapping parameters. The trapping process determines how far away an
edge is from the selected pixel and the amount of the trapping to be
applied to the center pixel. The trapped value of the selected pixel
will be a linear interpolation of the two different color values
that form the edge. This process repeats for every color plane. For
the ease of hardware implementation, all the above seven steps are
repeated for every color plane. However, the parameters obtained
from the previous six steps do not change once they are calculated
for any one of the four planes. In our software implementation of
JTHBCT03 for running time comparison (see Table \ref{table: overall
time comparison}), we did not repeat all seven steps for every color
plane, but rather only the last one. The rest of JTHBCT03 was
implemented as described in~\cite{Tr03}.

%% file: figs/tolerance_vol.pstex_t
\begin{picture}(0,0)%
\includegraphics{tolerance_vol.pstex}%
\end{picture}%
\setlength{\unitlength}{3947sp}%
\begingroup\makeatletter\ifx\SetFigFont\undefined%
\gdef\SetFigFont#1#2#3#4#5{%
  \reset@font\fontsize{#1}{#2pt}%
  \fontfamily{#3}\fontseries{#4}\fontshape{#5}%
  \selectfont}%
\fi\endgroup%
\begin{picture}(4438,3162)(1876,-4111)
\put(2701,-1711){\makebox(0,0)[lb]{\smash{{\SetFigFont{12}{14.4}{\rmdefault}{\mddefault}{\updefault}{\color[rgb]{0,0,0}$K_{A_{hi}}$}%
}}}}
\put(2701,-2986){\makebox(0,0)[lb]{\smash{{\SetFigFont{12}{14.4}{\rmdefault}{\mddefault}{\updefault}{\color[rgb]{0,0,0}$K_{A_{lo}}$}%
}}}}
\put(5176,-2536){\makebox(0,0)[lb]{\smash{{\SetFigFont{12}{14.4}{\rmdefault}{\mddefault}{\updefault}{\color[rgb]{0,0,0}$C_{A_{hi}}$}%
}}}}
\put(4576,-3136){\makebox(0,0)[lb]{\smash{{\SetFigFont{12}{14.4}{\rmdefault}{\mddefault}{\updefault}{\color[rgb]{0,0,0}$C_{A_{lo}}$}%
}}}}
\put(2851,-3286){\makebox(0,0)[lb]{\smash{{\SetFigFont{12}{14.4}{\rmdefault}{\mddefault}{\updefault}{\color[rgb]{0,0,0}$M_{A_{lo}}$}%
}}}}
\put(4201,-3286){\makebox(0,0)[lb]{\smash{{\SetFigFont{12}{14.4}{\rmdefault}{\mddefault}{\updefault}{\color[rgb]{0,0,0}$M_{A_{hi}}$}%
}}}}
\end{picture}%

%% file: LUT-based.tex
\section{LUT-based pixel-dependent color trapping}
\label{LUT approaches}

JTHBCT03 is very efficient when implemented in hardware but not in
software. In this section, we propose two software implementations
of this algorithm where all the algebraic operations are replaced by
LUTs in order to improve the computational speed. First, the steps
of color categorization, color density calculation, and trapping
parameter calculation can all be replaced by small LUTs. The feature
extraction and the edge detection steps, where a $5\times5$ $A$,
$B$, and $O$ color arrangement is classified into one of three
categories ({\em edge1}, {\em edge2}, or {\em non-trappable}), can
also be replaced by LUTs. This is one approach we propose, which
will be further discussed in Sec.~\ref{5x5 trapping}. However,
because of the size of the window considered, this approach has a
relatively large memory requirement. One way to reduce the memory
required consists of reducing the size of the window considered to
$3\times3$ pixels. Taking into consideration all the possible pixel
configurations within the $3\times3$ window, one can derive the edge
classification table based on the edge rules developed in JTHBCT03.
Since the information about the outer ring statistics is missing
from the $3\times3$ window, this approach is not so effective as the
full $5\times5$ window approach in correcting all possible gap or
halo artifacts. However, it can still efficiently eliminate the
appearance of white gaps next to black areas as well as some gap or
halo artifacts due to other colors caused by misregistration of up
to one pixel in extent. We now present the details of our proposed
implementations of the $3\times3$ and the $5\times5$ sliding window
approaches. We call these implementations {\em pixel-dependent}, as
the data related to overlapping pixels between two adjacent windows
are stored in a buffer to further reduce the computation time, as
will be explained shortly.

\subsection{Algorithm 1: Pixel-Dependent LUT-Based Trapping with a $3\times3$ Sliding Window}
\label{subsec:3x3 trapping}

\begin{figure}[!t]
\centerline{ \scalebox{0.8}{
\input{./figs/ktol.pstex_t} }}
\caption{Calculation of $K_{TOL}$ as a piecewise linear function of
the $K$ value of the center pixel of the evaluated $3\times3$
window, $K_{A}$. According to the Weber-Fechner law, the size of the smallest
noticeable difference of Human Visual System (HVS) is roughly
proportional to the intensity of the stimulus. Therefore, as $K$
value increases, more colors can be grouped into one type without
causing noticeable difference.}\label{fig:ktol}
\end{figure}
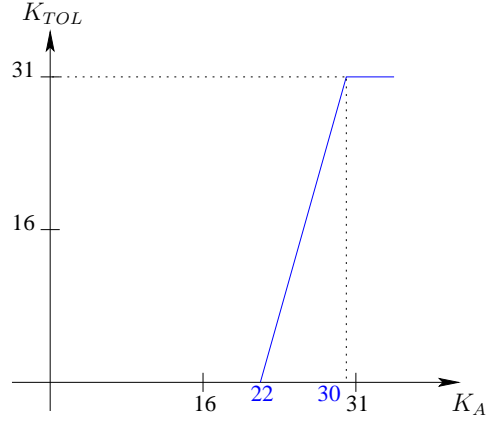

The first step of our implementation which differs from that of
JTHBCT03 is the color categorization (Step Two), in which a
parameter $K_{TOL}$ that represents the tolerance level as
determined by the $K$ value $K_{A}$ of the center pixel is computed.
The function representing the relationship between $K_{TOL}$ and
$K_{A}$, shown in Fig.~\ref{fig:ktol}, depends on the parameter
$A_{TOL}$, which specifies the minimum size of the volume in which
colors are similar and can be grouped into one. Following JTHBCT03,
we are using the value $A_{TOL}=24$. Once the $K_{TOL}$ value is
determined, the lower and the upper bound of the tolerance volumes
($X_{A_{lo}}$ and $X_{A_{hi}}$ for $A$; $X_{B_{lo}}$ and
$X_{B_{hi}}$ for $B$) are computed based on the $C$, $M$, and $K$
values of the center pixel. Note that since color $K$ has more
impact on the perceived color of the pixel than $C$ and $M$ do, the
larger the $K$ value is, the larger the tolerance volume. Finally,
the limits of the tolerance volume are shifted such that all the
color pixel values are guaranteed to range from $0$ to $255$. More
precisely, the tolerance volume of color $A$ pixels can be computed
as follows:

\setlength{\arraycolsep}{0.14em}{
\begin{eqnarray}
\label{eq:atol}
A_{TOL} & = & 24,\\
\label{eq:btol}
B_{TOL} & = & \max(A_{TOL}, K_{TOL}),\\
\label{eq:xalo}
X_{A_{lo}} & = &
    \left\{
    {
     \begin{array}{*{20}l}
     {\min(\max(X-B_{TOL},0),255-2B_{TOL}),} & & {\text{if $X = C, M$,}} \\
     {\min(\max(X-A_{TOL},0),255-2A_{TOL}),} & & {\text{if $X = K$,}}    \\
     \end{array}
     }
    \right. \\
\label{eq:xahi}
X_{A_{hi}} & = &
    \left\{
    {
     \begin{array}{*{20}l}
     {\max(\min(X+B_{TOL},255),2B_{TOL}),} & & & {\text{if $X = C, M$,}} \\
     {\max(\min(X+A_{TOL},255),2A_{TOL}),} & & & {\text{if $X = K$.}}    \\
     \end{array}
    }
    \right.
\end{eqnarray}}

Similarly, the tolerance volume for color $B$ pixels can be computed
as
\setlength{\arraycolsep}{0.14em}{
\begin{eqnarray}
\label{eq:xblo}
X_{B_{lo}} & = &
\min(\max(X-B_{TOL},0),255-2B_{TOL}), \qquad \text{for $X = C, M, K$,} \\
\label{eq:xbhi} X_{B_{hi}} & = & \max(\min(X+B_{TOL},255),2B_{TOL}),
\qquad \text{for $X = C, M, K$.}
\end{eqnarray}}

However, since $K_{A}$ as well as the $C$, $M$, and $K$ pixel values
take on integer values within the range $0$ to $31$, we pre-compute
all the tolerance values $X_{hi}$ and $X_{lo}$ for every possible
$K_{A}$, and store them in a small LUT.

Instead of performing feature extraction (Step Three) and edge
detection (Step Four) as in JTHBCT03, we designed a LUT, where the
edge classification results are stored for all possible color pixel
configurations. To do this, we observe that for a trappable edge to
exist in a $5\times5$ window, JTHBCT03 assumes that there are not
too many distinct colors inside the inner ring of the window
(Fig.~\ref{fig:scan_order}). In particular, if a color $O$ pixel is
present in the inner ring, then the center pixel can immediately be
declared to be {\em non-trappable}. In other words, we only need to
worry about windows that contain no color $O$ pixels in the inner
ring. Since we are using a window of size $3\times3$, this means
that we only need to consider windows containing only color $A$ and
$B$ pixels. We use bit `$1$' to denote $B$ pixels, and `$0$' to
denote $A$ pixels, thus obtaining a simple 8-bit label for each
window configuration to be classified. (Note that by definition the
center pixel is always an $A$ pixel; so only eight out of nine
binary pixel values need to be used to index into the LUT.) The
corresponding classification, either {\em edge1} or {\em
non-trappable}, is found using Trask's edge rules and stored in the
LUT. Note that the {\em edge2} type is not used here, as the window
size is too small to detect pixels that are one pixel away from an
edge. The structure of the LUT is illustrated in Fig.~\ref{fig:3x3
lut}. Our implementation of the trapping process leading to the edge
detection is summarized in Fig. \ref{fig:3x3 edge detection}.

\begin{figure}[!t]
\centerline{\epsfig{figure=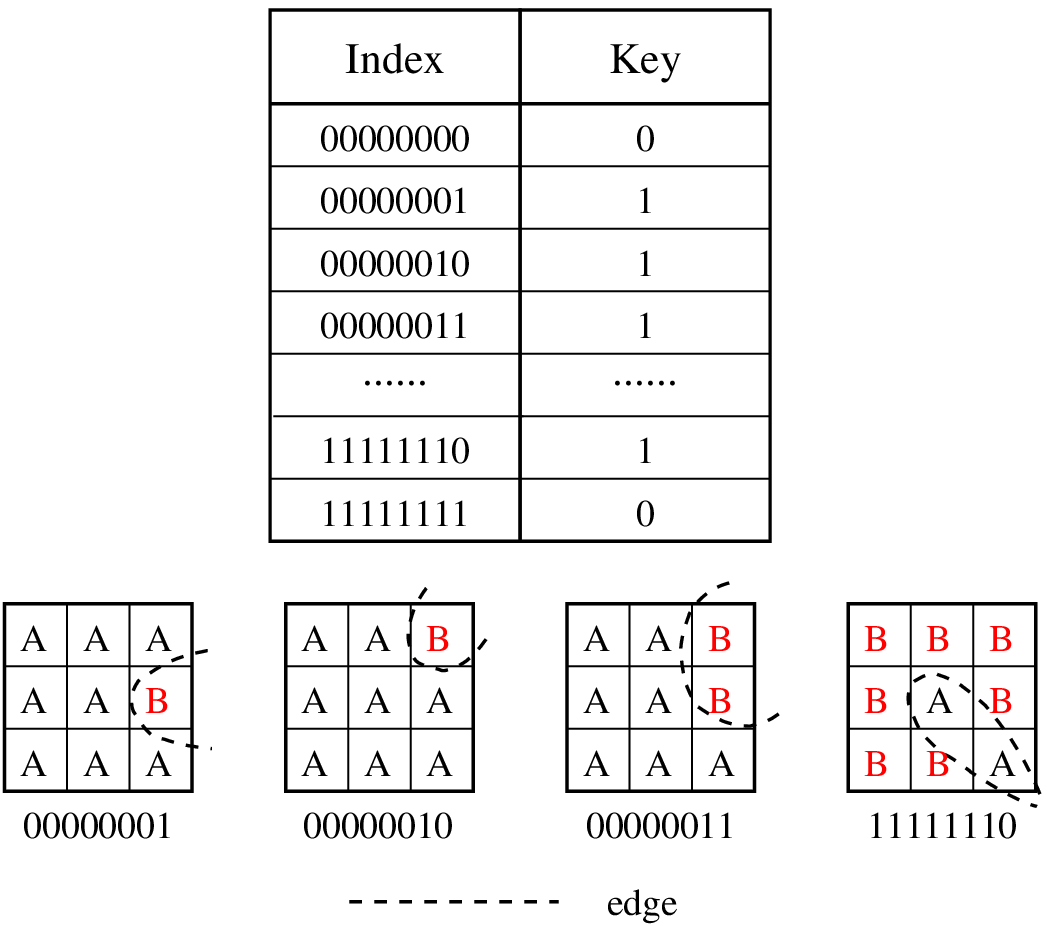, height = 3in}}
\caption{The look-up table for the edge detection using a $3\times3$
sliding window. Every color pixel arrangement in the window is used
as an index to access the table entry, which represents whether the
evaluated pixel is non-trappable ($0$) or trappable ($1$).
}\label{fig:3x3 lut}
\end{figure}

\begin{figure}[!t]
\begin{center}
$\begin{array}{cc}
    \scalebox{0.8}{
    \input{./figs/3x3_ed_diagram.pstex_t}} & \\[-.05cm]
    \mbox{(a)}
\end{array}$
\end{center}
\begin{center}
$\begin{array}{cccc}
    \epsfig{file=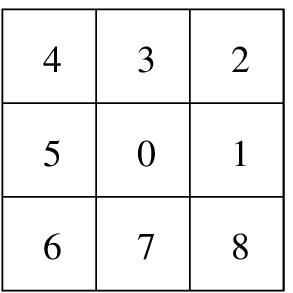, width = .8in} & & &
    \epsfig{file=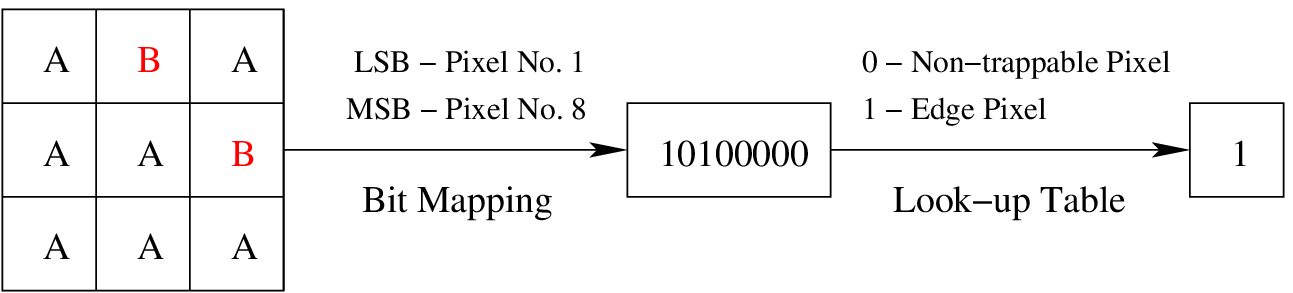, height= .8in} \\[-.05cm]
    \mbox{(b)} & & & \mbox{(c)}
    \end{array}$
\end{center}
{\vspace{-0.1cm}} \caption{LUT-based edge detector using a
$3\times3$ sliding window. (a) Block diagram. (b) Pixel ordering for
the $3\times3$ window. (c) An example of edge detection process. Bit
'$0$' represents a color $A$ pixel, and bit '$1$' represents a color
$B$ pixel. The color types of the pixels in the window are mapped to
a binary representation in a certain order so that the pixel no. $1$
is mapped to the LSB (Least Significant Bit) and pixel no. $8$ is
mapped to the MSB (Most Significant Bit).} \label{fig:3x3 edge
detection}
\end{figure}


Once the evaluated pixel is determined to be trappable ({\em
edge1}), it needs to be trapped. In our implementation, the
applicable trapping parameters that control the amount of edge
movement, if any, are retrieved from a LUT. They are pre-computed
from the piecewise linear functions (with discrete values as inputs
and outputs) used in JTHBCT03. A description of these functions can
be found in~\cite{Tr03}. Based on the trapping parameters, the final
trapped color values are also obtained from a LUT rather than
algebraic equations.

We observe that both the tolerance values for color $A$ pixels and
most of the trapping parameters (see~\cite{Tr03}) are determined by
the color values of the center pixel. Therefore, if two neighboring
pixels are of the same color, then as the sliding window moves from
the position where the one on the left is the center pixel in the
window to the position where the one on the right becomes the center
pixel in the window, the values of those parameters remain
unchanged. This observation can be used to reduce the number of
times each pixel needs to be visited. Since pairs of similar color
pixels are often found next to each other, a significantly shorter
running time can be achieved as a result. The implementation of this
pixel-dependent approach is illustrated in Fig.~\ref{fig:3x3 flow
diagram}.

\begin{figure}[tbp]
\centerline{\epsfig{figure=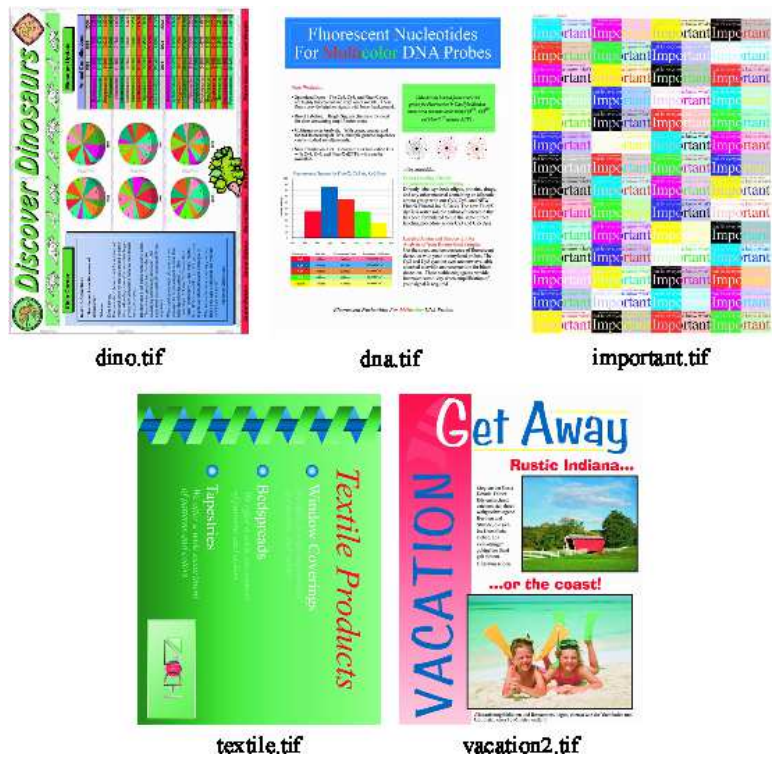, width=4in}}
\caption{Thumbnails of sample test images used to evaluate the
performance of our three color trapping algorithms. All images are
$C$, $M$, $Y$, $K$ ($32$ bits/pixel), and are of size $6,400\times
4,900$ pixels, corresponding to a $600$ dpi rendering on an $8.5
\times 11$ inch page.}\label{fig:test images}
\end{figure}

The running time comparison between the pixel-dependent and
pixel-independent approaches is shown in Table \ref{tb:3x3 running
time comparison}. The tests were performed on a machine with an
Intel\textregistered \textrm{ }Xeon(TM) processor and CPU speed of
$3.60$ GHz using $5$ images: $dino.tif$, $dna.tif$, $important.tif$,
$textile.tif$, and $vacation2.tif$ which are shown in
Fig.~\ref{fig:test images}. The algorithm was run $10$ times on each
image and the average running time was recorded. For the $5$ images,
the average speed gain of the pixel-dependent approach over the
pixel-independent approach is above $33\%$. As can be seen from
Table \ref{table: overall time comparison}, both of the LUT-based
$3\times3$ sliding window approaches are significantly faster than
JTHBCT03.

\begin{figure}[!t]
\centerline{ $\begin{array}{cc}
    \epsfig{file=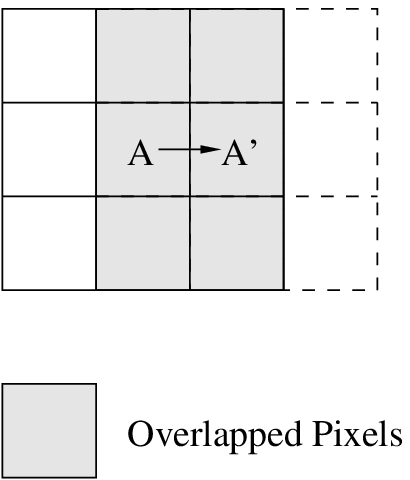, height = 1in} \\[-.05cm]
    \mbox{(a)}
\end{array}$
$\begin{array}{cc}
    \epsfig{figure=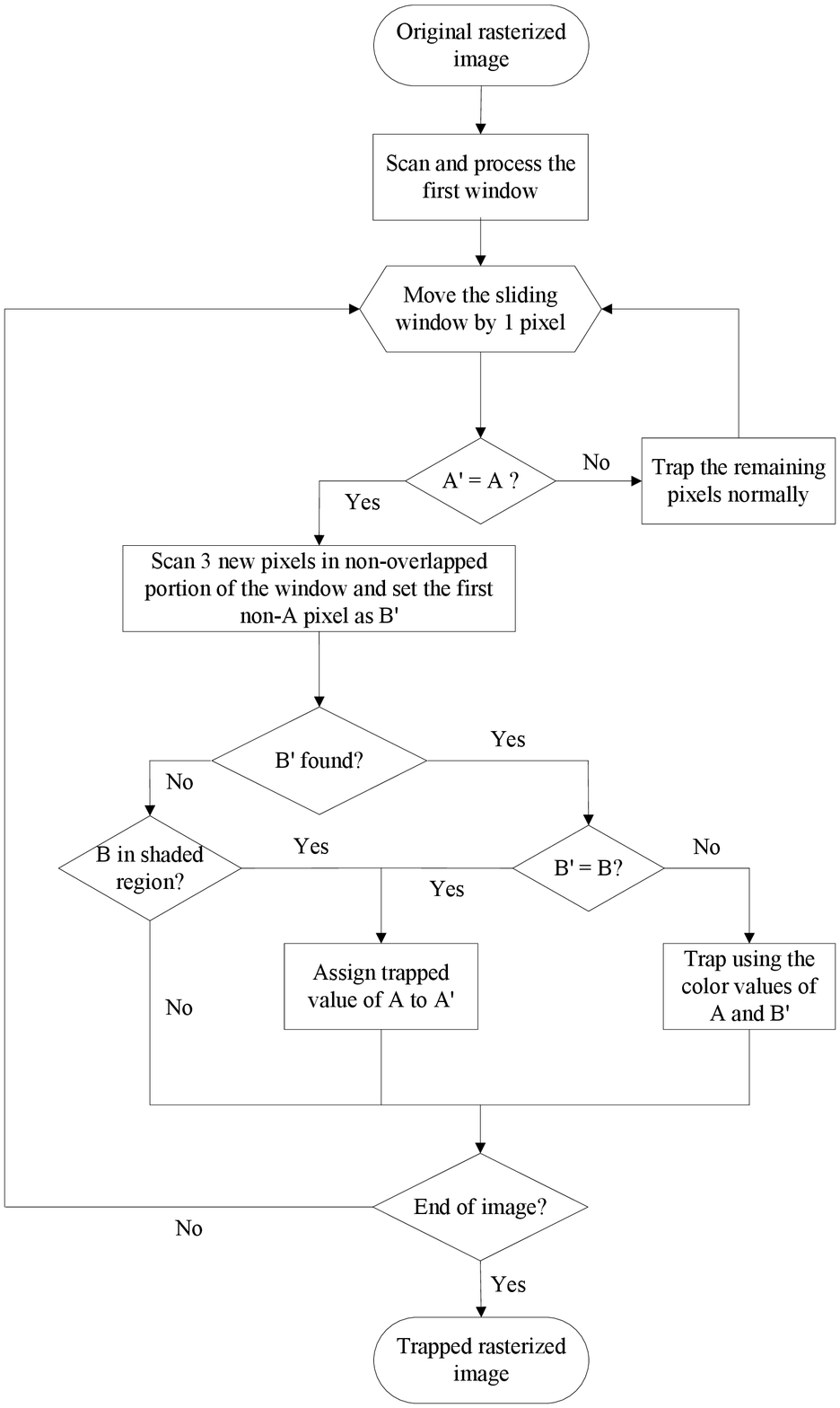, height = 5in} \\[-.05cm]
    \mbox{(b)}
\end{array}$}
\caption{Flow diagram of LUT-based pixel-dependent color trapping
approach using a $3\times3$ sliding window (Algorithm 1). The center
pixel of the window is called $A$. The first pixel not in the same
color group as $A$ is set to $B$. When the window moves, the new
center pixel is called $A'$. (a) Movement of a $3\times3$ sliding
window by one pixel. (b) Flow diagram for the pixel-dependent
approach. }\label{fig:3x3 flow diagram}
\end{figure}

\begin{figure}[!tbp]
\begin{center}
\epsfig{figure=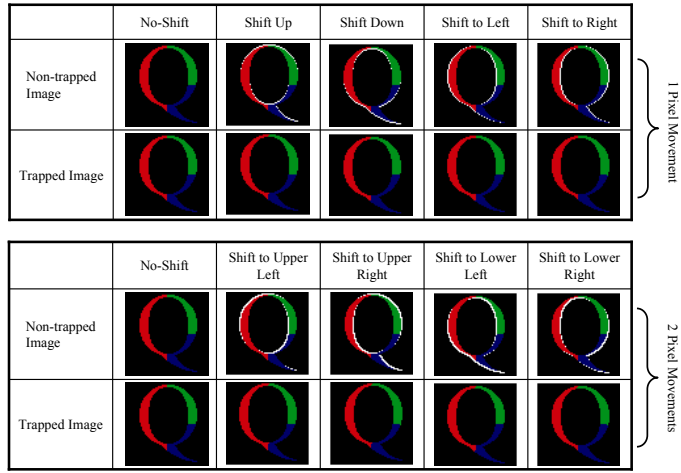, width = 3in}
\end{center}
\caption{Test results for an image with $K$ plane misregistration
using a $3\times3$ sliding window. The black plane was shifted by
all combinations of $1$ pixel up, down, left, and right to simulate
the effect of printer misregistration.}\label{fig:3x3 test results}
\end{figure}

Experimental results on an image with $K$ plane misregistration are
shown in Fig. \ref{fig:3x3 test results}. As can be seen, our
pixel-dependent color trapping approach using LUTs can effectively
eliminate the white gap next to black when the $K$ plane is
mis-aligned by up to one pixel in extent in either horizontal or
vertical directions or both. The memory cost added by the LUTs is
$3794$ bytes (see Table \ref{tb:3x3 memory requirement}). Assuming
we store $5$ lines of a full page image ($6400\times4900$) in the
memory at a time, the total memory required to store the image data,
the LUTs, dummy variables, and all the trapping parameters is around
$727$ Kbytes.

\begin{table*}[tbp]
\renewcommand{\arraystretch}{1.3}
\begin{center}
\caption{Running time comparison of pixel-independent and
pixel-dependent approaches for LUT-based trapping with a $3\times3$
sliding window (Algorithm 1).} \label{tb:3x3 running time
comparison}
\begin{minipage}{\linewidth}
\centering
\renewcommand{\thefootnote}{\thempfootnote}
\begin{tabular}{|c||c|c|}
\hline \hline \multirow{2}{*}{File Name\footnote{All the test images
are of size $6400\times 4900$.} ($.tif$)} &
\multicolumn{2}{c|}{Running Time\footnote{The tests were performed
on a linux machine with an Intel\textregistered \textrm{ }Xeon(TM)
processor and CPU speed of $3.60$ GHz.} (sec)} \\[0.5ex]
\cline{2-3}
& Pixel-independent Approach & Pixel-dependent Approach \\
\hline 

$dino$ & $3.4$ & $2.29$ \\

$dna$ & $2.46$ & $1.76$ \\

$important$ & $4.03$ & $2.62$ \\

$textile$ & $4.44$ & $2.85$ \\

$vacation2$ & $3.71$ & $2.53$ \\

\hline

Average & $3.608$ & $2.41$ \\

\hline
\end{tabular}
\end{minipage}
\end{center}
\end{table*}

\begin{table}[!t]
\renewcommand{\arraystretch}{1.3}
\begin{center}
\caption{Memory required for LUT-based trapping with a
\mbox{$3\times3$} sliding window (Algorithm 1).} \label{tb:3x3
memory requirement}
\begin{minipage}{\linewidth}
\centering
\renewcommand{\thefootnote}{\thempfootnote}
\begin{tabular}{|c|c|}
\hline \hline Data To Be Stored & Memory Required (bytes)\\
\hline
$K_{TOL}$ & $32$ \\
$X_{lo}$ & $1,024$ \\
$X_{hi}$ & $1,024$ \\
Edge Detection LUT & $256$ \\
Trapping Parameters\footnote{All the entries in the LUTs are stored
as unsigned characters of size $1$ byte each. The edge detection LUT
contains $256$ entries for the $256$ different color pixel
arrangements. The trapping parameters include $KDIF$, $OVRD$,
$TRAP$, and $EDGE$
(refer to \cite{Tr03}).} & $1,458$ \\
\hline
Subtotal & $3,794$ \\
\hline
$5$-Line Pixel Buffer & $98,000$ \\
\hline Other Variables Used\footnote{The other variables include all
the temporary variables (such as the edge type and the center pixel
color information)
used to execute the algorithm.} & $\approx642,654$ \\
\hline
Total & $\approx727K$ \\

\hline
\end{tabular}
\end{minipage}
\end{center}
\end{table}

\subsection{Algorithm 2: Pixel-Dependent LUT-Based Trapping with a $5\times5$ Sliding Window}
\label{5x5 trapping}

Our approach for the $5\times5$ sliding window case is analogous to
that of the $3\times 3$ window. The main difference is that there
are more pixels to take into account inside the window. While this
adds to the memory and the computational requirements, the resulting
trapping can effectively hide misregistration errors up to two
pixels in extent in either horizontal or vertical directions or
both.

\begin{figure}[!t]
  \centering
  \subfigure[No $O$ pixel.]{\label{sf:O0}\includegraphics[width=0.15\textwidth]{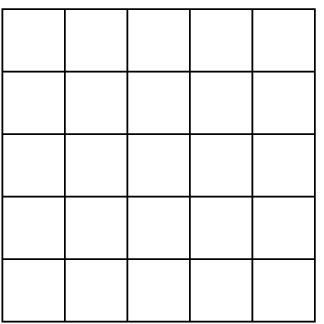}}\\
  \vspace{.1in}
  \subfigure[One $O$ pixel.]{\label{sf:O1}\includegraphics[width=0.3\textwidth]{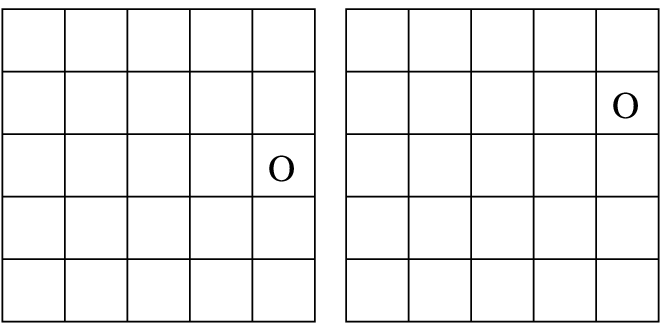}}
  \hspace{.1in}
  \subfigure[Two $O$ pixels.]{\label{sf:O2}\includegraphics[width=0.3\textwidth]{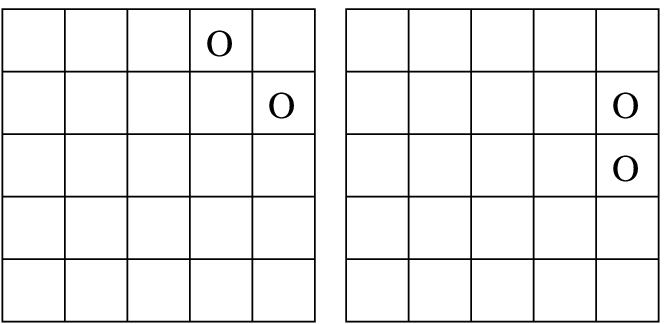}}\\
  \vspace{.1in}
  \subfigure[Three $O$ pixels.]{\label{sf:O3}\includegraphics[width=0.3\textwidth]{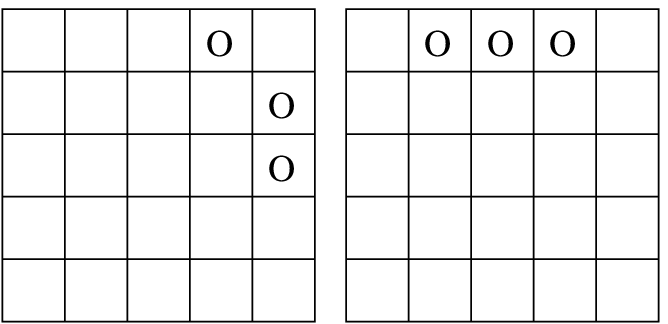}}
  \hspace{.1in}
  \subfigure[Four $O$ pixels.]{\label{sf:O4}\includegraphics[width=0.3\textwidth]{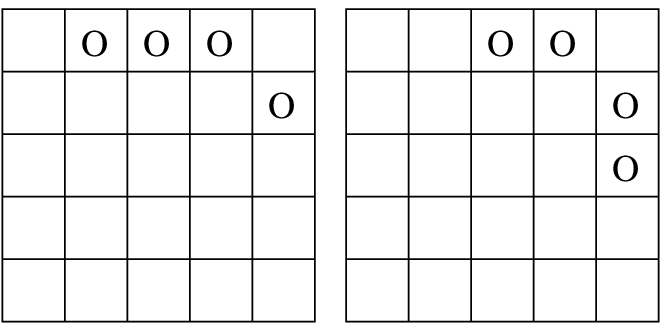}}\\
  \vspace{.1in}
  \subfigure[Five $O$ pixels.]{\label{sf:O5}\includegraphics[width=0.3\textwidth]{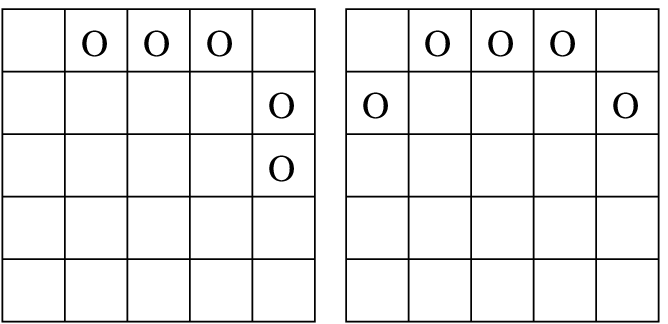}}
  \hspace{.1in}
  \subfigure[Six $O$ pixels.]{\label{sf:O6}\includegraphics[width=0.3\textwidth]{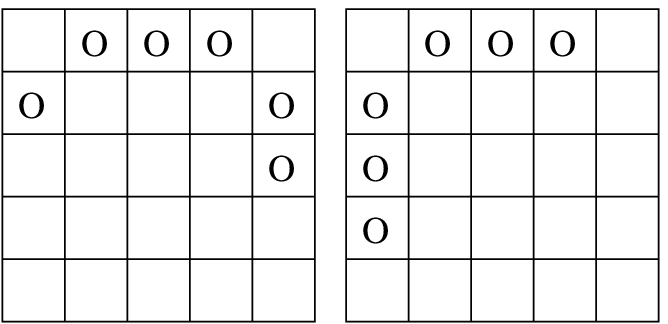}}\\
  \caption{Arrangements of $O$ pixels in the outer ring of the $5\times5$ window used for indexing into one of the edge classification LUTs of Algorithm 2. Note that corner pixels are excluded as shown in Fig.~\ref{fig:scan_order}.
  All the patterns which are not shown here are symmetric (under a flip and rotation) to one of the arrangements above.  The symmetric pattern correspondence is stored in another  LUT. }
  \label{fig:o arrangements}
\end{figure}

We implemented the color categorization step using a LUT in a
fashion similar to that described in Sec.~\ref{subsec:3x3 trapping}.
However, the edge detection step is more complicated in this case.
One reason for this is that there are many more ($3^{20}/2$)
different color pixel arrangements. Moreover, the existence of $O$
pixels in the outer ring of the window creates a third color label
which makes it impossible to directly obtain a label using the
binary scheme of the $3\times 3$ case. Instead, a set of LUTs (Fig.
\ref{fig:o arrangements}) were designed based on the number and
positions of $O$ pixels in the outer ring of the window.

Since there are $3^{20}/2$ different pixel combinations, it is not
feasible to use them directly to index into a LUT. Instead, the
window is flipped and rotated such that the number and positions of
$O$ pixels identified in the outer ring of the $5\times5$ window
match one of the thirteen cases shown in Fig.~\ref{fig:o
arrangements}. This step is valid since the edge type is invariant
with respect to a rotation and a flip of the window. The flipping
and rotation operations are also designed using LUTs.

The color trapping step is implemented in a way similar to that for
a $3\times3$ window. Again, we observe that in consecutive
positions, the windows have $21$ overlapping pixels which leads to
the implementation of a faster, pixel-dependent approach. But since
there are more overlapping pixels than when using the smaller
($3\times3$) window, the running time can be further reduced for
images containing regions of uniform colors. The average running
time comparisons between JTHBCT03 \cite{Tr03}, our pixel-independent
LUT-based approach, and our pixel-dependent LUT-based approach are
tabulated in Table \ref{tb:5x5 running time comparison}. As can be
observed, by making use of the information for the overlapping
pixels and LUTs, the pixel-dependent LUT-based approach is about
$25\%$ faster than the pixel-independent approach, and more than
three times faster than JTHBCT03.

\begin{table}[!t]
\renewcommand{\arraystretch}{1.3}
\begin{center}
\caption{Running time comparison of pixel-independent and
pixel-dependent approaches for LUT-based trapping using a $5\times5$
sliding window (Algorithm 2).} \label{tb:5x5 running time
comparison}
\begin{minipage}{\linewidth}
\centering
\renewcommand{\thefootnote}{\thempfootnote}
\begin{tabular}{|c|c|}
\hline \hline Approach Taken & Average Running Time\footnote{The
running time tests were performed on a machine with an
Intel\textregistered \textrm{ }Xeon(TM) processor and CPU speed of
$3.60$ GHz for $5$ images $dino.tif$, $dna.tif$, $important.tif$,
$textile.tif$, and $vacation2.tif$ which are shown in Fig.
\ref{fig:test images}. The running time is taken as an average of
the average running time for the $5$ test images, each
of which was tested $10$ times.} (secs)\\

\hline
JTHBCT03 & $25.83$ \\

Pixel-independent & $9.05$ \\

Pixel-dependent (Algorithm 2) & $6.81$ \\

\hline
\end{tabular}
\end{minipage}
\end{center}
\end{table}

The memory required for processing a $5\times5$ window of pixel data
is tabulated in Table \ref{tb:5x5 memory requirement}. The LUTs
require approximately $3$ Mbytes of storage space. The total memory
requirement for this approach, including dummy variables, image data
($5$ lines of a full  $6,400\times 4,900$ pixel page) and all the
trapping parameters is around $3.7$  Mbytes.

\begin{table*}[!t]
\renewcommand{\arraystretch}{1.3}
\begin{center}
\caption{Memory required for LUT-based  trapping with a \mbox{$5\times5$} sliding window (Algorithm
$2$).} \label{tb:5x5 memory requirement}
\begin{minipage}{\linewidth}
\centering
\renewcommand{\thefootnote}{\thempfootnote}
\begin{tabular}{|c|c|c|}
\hline \hline
\multicolumn{2}{|c|}{Data To Be Stored} & Memory Required (bytes) \\
\hline

\multicolumn{2}{|c|}{$K_{TOL}$} & $32$ \\

\multicolumn{2}{|c|}{$X_{lo}$} & $1,024$ \\

\multicolumn{2}{|c|}{$X_{hi}$} & $1,024$ \\

\cline{1-2}
\multirow{7}{*}{LUTs for Edge Detection} & No $O$ in OR\footnote{Outer Ring} & $2^{20}$ \\[0.5ex]

& One $O$ in OR\footnotemark[\value{mpfootnote}] & $2^{20}$ \\

& Two $O$s in OR\footnotemark[\value{mpfootnote}] & $2^{19}$ \\

& Three $O$s in OR\footnotemark[\value{mpfootnote}] & $2^{18}$ \\

& Four $O$s in OR\footnotemark[\value{mpfootnote}] & $2^{17}$ \\

& Five $O$s in OR\footnotemark[\value{mpfootnote}] & $2^{16}$ \\

& Six $O$s in OR\footnotemark[\value{mpfootnote}] & $2^{15}$ \\

\cline{1-2}

\multicolumn{2}{|c|}{LUTs for flipping and rotation ($4$ flippings
and $3$ rotations)} & $147$ \\

\multicolumn{2}{|c|}{LUT for remapping edges} & $111$ \\

\multicolumn{2}{|c|}{LUTs for checking neighborhoods} & $40$ \\

\multicolumn{2}{|c|}{Trapping Parameters\footnote{All the entries in
the LUTs are stored as unsigned characters of size $1$ byte each.
Edge detection LUTs were designed based on the number and position
of $O$ pixels (Fig. \ref{fig:o arrangements}) in the outer ring of
the selected window. Trapping parameters include $KDIF$, $OVRD$,
$TRAP$ and $EDGE$ (refer
to~\cite{Tr03}).}} & $1,458$ \\
\hline

\multicolumn{2}{|c|}{Subtotal} &
$3,116,796\approx3M$ \\
\hline

\multicolumn{2}{|c|}{$5$-Line Pixel Buffer} & $98,000$ \\
\hline

\multicolumn{2}{|c|}{Other Variables Used} & $\approx642,654$ \\
\hline

\multicolumn{2}{|c|}{Total} & $3,857,450\approx3.76M$ \\
\hline

\end{tabular}
\end{minipage}
\end{center}
\end{table*}

%% file: figs/ktol.pstex_t
\begin{picture}(0,0)%
\includegraphics{KTOL.pstex}%
\end{picture}%
\setlength{\unitlength}{3947sp}%
\begingroup\makeatletter\ifx\SetFigFont\undefined%
\gdef\SetFigFont#1#2#3#4#5{%
  \reset@font\fontsize{#1}{#2pt}%
  \fontfamily{#3}\fontseries{#4}\fontshape{#5}%
  \selectfont}%
\fi\endgroup%
\begin{picture}(4200,3289)(4489,-3944)
\put(7951,-3886){\makebox(0,0)[lb]{\smash{{\SetFigFont{12}{14.4}{\rmdefault}{\mddefault}{\updefault}{\color[rgb]{0,0,0}$K_{A}$}%
}}}}
\put(4576,-811){\makebox(0,0)[lb]{\smash{{\SetFigFont{12}{14.4}{\rmdefault}{\mddefault}{\updefault}{\color[rgb]{0,0,0}$K_{TOL}$}%
}}}}
\end{picture}%

%% file: figs/3x3_ed_diagram.pstex_t
\begin{picture}(0,0)%
\includegraphics{3x3_ed_diagram.pstex}%
\end{picture}%
\setlength{\unitlength}{3947sp}%
\begingroup\makeatletter\ifx\SetFigFont\undefined%
\gdef\SetFigFont#1#2#3#4#5{%
  \reset@font\fontsize{#1}{#2pt}%
  \fontfamily{#3}\fontseries{#4}\fontshape{#5}%
  \selectfont}%
\fi\endgroup%
\begin{picture}(9709,897)(6204,-3973)
\put(6301,-3548){\makebox(0,0)[lb]{\smash{{\SetFigFont{12}{14.4}{\rmdefault}{\mddefault}{\updefault}{\color[rgb]{0,0,0}3$\times$3 Window}%
}}}}
\end{picture}%

%% file: hybrid.tex
\section{Algorithm 3: Hybrid Method for Color Trapping with a $5\times 5$ Sliding Window}
\label{hybrid method}

\begin{figure*}[!tbp]
\centerline{\epsfig{figure=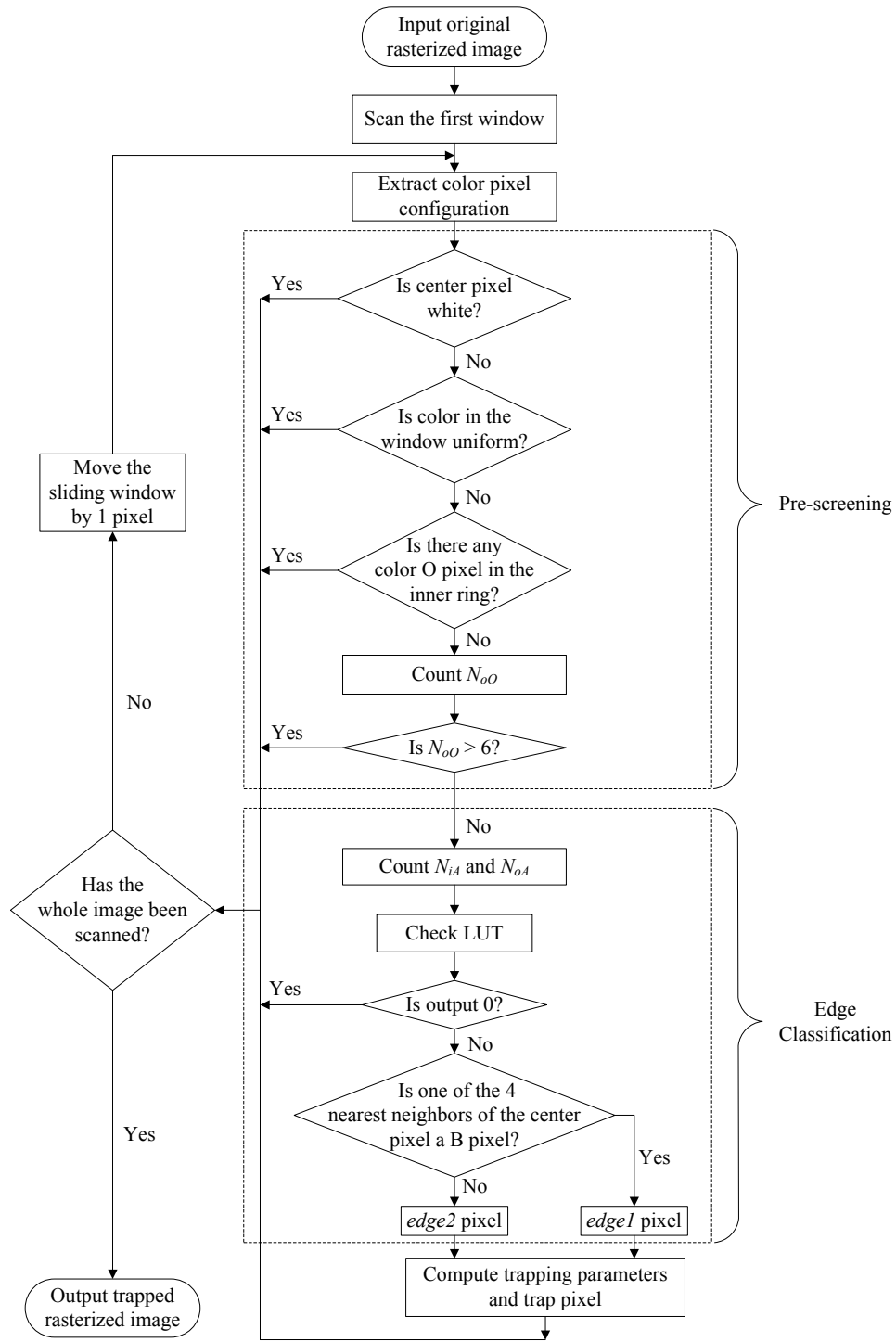, width = 5in}}
\caption{Flow diagram of proposed hybrid method for color trapping
(Algorithm 3), including the details of prescreening and edge
classification steps.}\label{fig:edge_detector}
\end{figure*}

Replacing computations by LUTs, as we did in the previous section,
is a well known way to reduce the complexity of an algorithm.
However, it is important to keep in mind that while the use of
look-up tables is often effective in gaining speed, memory is
required to store this information. Moreover, since accessing memory
also takes time, and since the access time may grow to some extent
with the size of the LUT, converting computations into LUTs is
sometimes not worthwhile from a CPU time perspective, especially
when the computations they replace are quite simple.

The LUTs used in the previous section to replace the steps of color
categorization, color density calculation, and trapping parameter
calculation are small enough to yield an obvious speed gain.
However, the LUT needed in Sec.~\ref{5x5 trapping} to replace the
feature extraction and the edge detection steps, where the
$5\times5$  $A$, $B$, and $O$ color arrangement is classified into
one of three categories ({\em edge1}, {\em edge2}, or {\em
non-trappable}), requires a significant amount of storage space. In
fact, the $3$ Mbytes of storage required to implement this approach
would be unacceptable for a low-cost formatter-based color printer.
While the $3\times3$ window approach has a significantly smaller
storage requirement, it only provides an approximation to the edge
classification results yielded by JTHBCT03. In this section, we
present an alternative approach combining feature selection and the
use of LUTs, which greatly reduces the storage space required
without significantly increasing the computation time. This approach
was developed using a training set consisting of $73$ typical print
images with sizes ranging from $4,200\times 2,925$ pixels to
$9,917\times 7,007$ pixels. This set of images contains various
types of text, graphics and pictures.

The flow diagram of our hybrid approach to classify edges is shown
in Fig. \ref{fig:edge_detector}. This hybrid approach combines
feature extraction and a $3$-dimensional LUT to achieve fast and
efficient edge classification. The first part of this approach
consists of a prescreening step which identifies the most frequent
and easily recognizable non-trappable color patterns using some
simple features. This step is followed by a slightly more
computationally expensive edge classification step consisting of
extracting three integer valued features to index into a LUT
separating trappable pixels ({\em edge1} and {\em edge2}) from
non-trappable pixels. A simple rule is then derived to distinguish
{\em edge1} from {\em edge2} pixels. We now describe both the
prescreening step and the edge classification step in greater
details.

\subsection{Prescreening}
\label{trask approach with prescreening}

One way to drastically reduce the computation time is to implement
some prescreening rules that quickly identify a large amount of
non-trappable pixels. By analyzing JTHBCT03, we observe the
following simple prescreening rules.
\begin{displaymath}
X = \left\{
\begin{array}{llll}
\textrm{{\em non-trappable}}, & & & \textrm{if X is white},\\
\textrm{} & & & \textrm{if the region has uniform color},\\
\textrm{} & & & \textrm{if number of inner $O$ pixels $>0$},\\
\textrm{} & & & \textrm{if number of outer $O$ pixels $>6$},\\
\textrm{needs further steps to detect its edge type}, & & &
\textrm{else},
\end{array}
\right.
\end{displaymath}
where $X$ represents the center pixel in the $5\times5$ window.

Here are short explanations for each of these prescreening criteria.

\subsubsection{White Pixel}
\label{white pixel} In order to preserve the fidelity of the image
and the outline of objects, white color pixels should remain
unchanged. Therefore, if the center pixel of the $5\times5$ window
is white ($C=0$, $M=0$, $Y=0$, and $K=0$), this pixel can be ignored
by the trapping algorithm.

\subsubsection{Uniform Color Region}
\label{uniform color region} Since gap or halo artifacts only occur
around the edge areas, uniform regions, i.e., where all the pixels
in the $5\times5$ window have the same color, should not be trapped.

\subsubsection{Number of Inner O Pixels}
\label{number of inner o pixels} If there is an $O$ pixel in the
inner ring of the $5\times5$ window, it is more likely that the
window is a part of a picture instead of a text or graphic object.
Therefore, pixels surrounded by such a color configuration should
not be trapped.

\subsubsection{Number of Outer O Pixels}
\label{number of outer o pixels} Gap or halo artifacts are more
visible in images containing text or graphics with only a few
density levels and sharp color transitions. For pictures with smooth
color changes, there is no need for trapping. As the presence of
many color $O$ pixels suggests that the window is likely to be part
of a picture, the center pixel of a window with more than six $O$'s
in the outer ring is not trapped.

\subsection{Edge Classification}
\label{sub:hybrid-based edge classification}

Any $5\times5$ window configuration that is not declared {\em
non-trappable} at the prescreening step needs to be further analyzed
in order to be classified into an edge category ({\em edge1}, {\em
edge2}, or {\em non-trappable}). JTHBCT03 necessitates the
extraction of $27$ discrete features in order to do this. Our aim is
to classify the pixels as accurately as possible with as few and
simple features as possible.

An important aspect of feature selection is to analyze the feature
space distribution. Obviously, the distribution of the color values
of the pixels is not uniform. That is, some color combinations occur
more frequently than others in actual images. We used our training
set to characterize the feature space distributions and the edge
rules of JTHBCT03 to establish the ground truth for the edge type.
We thoroughly analyzed all $27$ features used in JTHBC03 along with
four additional ones, namely the amount of scatter (i.e.~the
principal values) along the principal directions of the color A and
color B pixels, respectively. More precisely, we considered all
possible choices of $N$ features, for $N=1,2,\ldots,6$, and computed
a decision surface between the edge pixels and the non-edge pixels
inside the corresponding feature space using a support vector
machine algorithm. We chose to use no more than 6 features in order
to limit the computational requirements of the algorithm. Note that
we specifically did not use a dimension reduction technique such as
principal component analysis (PCA) to reduce the dimensionality of
the feature space, as our objective was to limit the number of
features we would have to compute to a strict minimum. The principal
values of the color A pixel arrangement were found to be quite
useful in determining the presence of an edge. However, we found
that the additional number of non-trappable edges identified using
these features did not outweigh the computational cost of extracting
these features and computing the value of the decision function. We
found that extracting discrete features and using them to index into
a LUT containing edge types was a more efficient approach. In fact,
we found that only three simple features are needed to simulate
JTHBCT03's edge classification on non-trappable and trappable pixels
with good accuracy. These are:
\begin{enumerate}
\item  $N_{oA}$ -- the number of $A$ pixels in the outer ring of the $5\times5$ window;
\item $N_{iA}$ -- the number of $A$ pixels in the $3\times3$ inner window of the $5\times5$ window;
\item $N_{oO}$ -- the  number of $O$ pixels in the outer ring of the $5\times5$ window.
\end{enumerate}

\begin{figure}[!t]
\centerline{\epsfig{figure=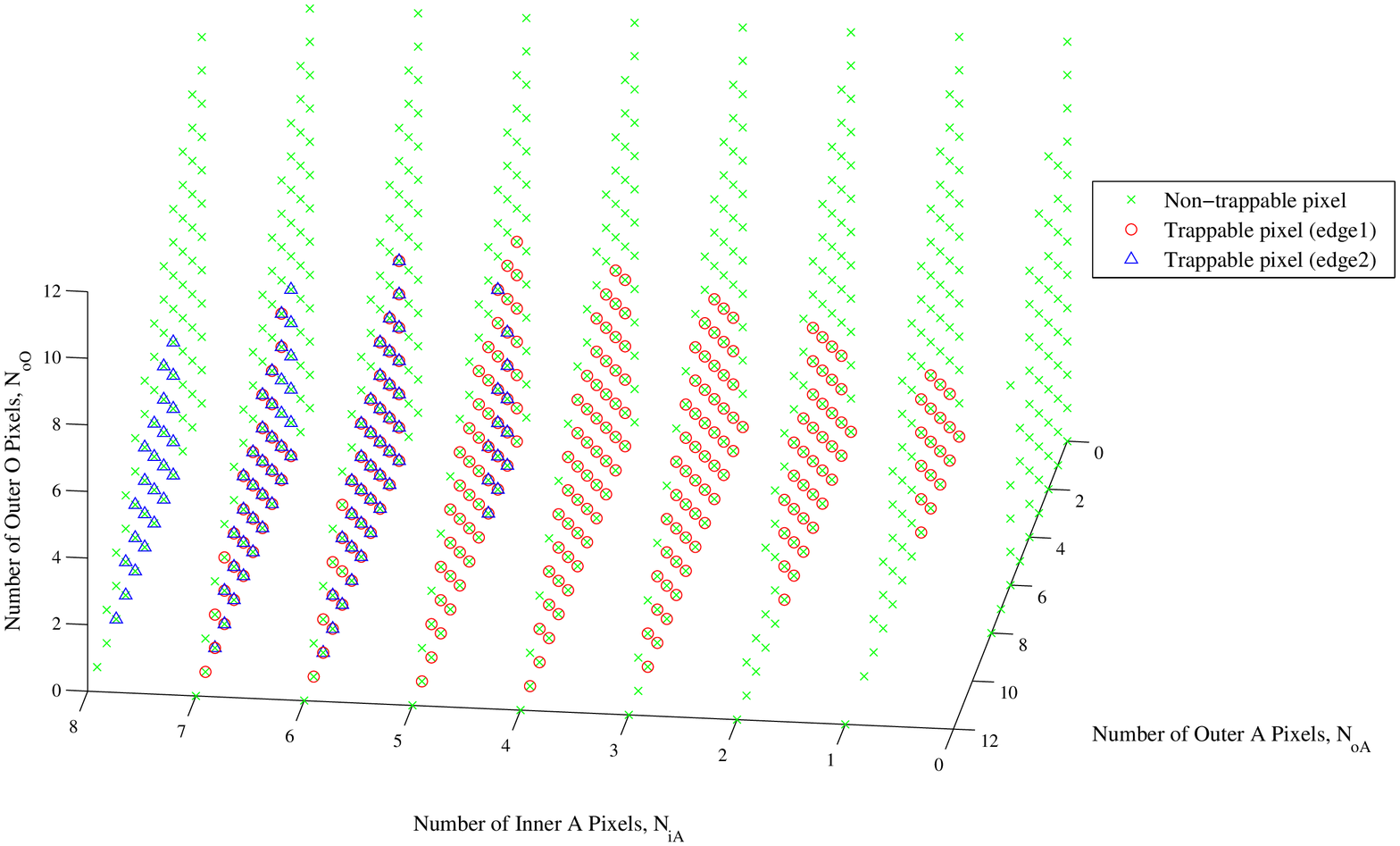, width = 6in}}
\caption{The $3$-dimensional plot for the number of inner \mbox{$A$}
pixels \mbox{$N_{iA}$}, the number of outer $A$ pixels
\mbox{$N_{oA}$}, and the number of outer $O$ pixels \mbox{$N_{oO}$}
in the selected window.} \label{fig:3-dimensionalplot}
\end{figure}

The resulting $3$-dimensional feature plot is shown in
Fig.~\ref{fig:3-dimensionalplot}. One notices that some points in
the feature space correspond to more than one edge type. We resolve
such ambiguities as follows: if JTHBCT03 considers all
configurations with a given value of $N_{oA}$, $N_{iA}$, and
$N_{oO}$ to be {\em non-trappable}, then we declare them to be {\em
non-trappable} as well; otherwise we trap them. Each possible
three-tuple of these features is indexed into a LUT where the
corresponding edge type is stored. Note that using a decision
surface instead of a LUT would result in a much more complex
decision rule, as the separation between the trappable and
non-trappable pixels is not linear, as one can easily see from the
feature plot.

Once a trappable pixel has been identified, its edge type ({\em
edge1} or {\em edge2}) needs to be determined. To do this, we check
whether there is any color $B$ pixel in the $3\times3$ inner window
that is one of the $4$ nearest neighbors of the center pixel (i.e.,
pixels numbered $1$, $3$, $5$, and $7$ as shown in
Fig.~\ref{fig:scan_order}). If there is, then the pixel is
categorized as {\em edge1}, otherwise it is categorized as {\em
edge2}. This is nearly always consistent with JTHBCT03's rules.

\subsection{Numerical Experiments and Discussion}
\label{hybrid-results and discussion}

The computational burden added by the prescreening step is small
considering the high number of non-trappable patterns it identifies.
Indeed, our training set contains a total of $1,391,758$ (unique)
non-trappable patterns,  $1,057,779$ ($76\%$) of which are
identified at the prescreening stage. However, non-trappable pixels
occur statistically more often than trappable pixels. In fact, the
prescreening step effectively  filters out the majority
($\approx90\%$) of all pixels, thus greatly diminishing the
computational impact of the edge classification step.


Since our three-feature combination do not completely differentiate
trappable pixels from non-trappable ones, our edge classification
step sometimes makes mistakes. Indeed, as can be seen from
Fig.~\ref{fig:3-dimensionalplot}, some {\em non-trappable} pixels
(represented by green crosses) overlap with {\em edge1} (represented
by red circles) or {\em edge2} pixels (represented by blue
triangles) in our feature space. As explained earlier, such pixels
end up being trapped. Overall, after the prescreening step, only
$87,836$ (or $26.3\%$) of the $333,979$ unique patterns (i.e., not
counting repetitions) do not overlap with an {\em edge1} or {\em
edge2} pixel in the feature space, and are therefore declared {\em
non-trappable}. However, this decision is always correct. Moreover,
such patterns occur very often in images. In fact, in our training
images, this set ($26.3\%$) of patterns corresponds to $66.67\%$ of
all the non-trappable patterns entering the edge classification
step. The remaining patterns are all declared trappable, which is
oftentimes incorrect (in the sense that this is not the same
classification as produced by JTHBCT03). But considering that,
because of our highly efficient prescreening, only about $10\%$ of
the patterns enter the edge classification step, the overall error
rate is quite small, namely $1.53\%$ on average for our $5$ test
images.

From a computational point of view, our overall algorithm is quite
simple, as it requires very few operations per trapped pixel, in
comparison with JTHBCT03. The detailed counts for each operation for
JTHBCT03 and Algorithms $1$ and $2$ are given in
Table~\ref{tb:complexity analysis}.

\begin{figure}[!t]
  \centering
  \subfigure[Close-up view of a portion of the original image. The gray ($50\%$ $K$) letters on the color background may create a halo artifact if the colored planes are mis-registered. ]{\label{sf2:original}\includegraphics[width=0.8\textwidth]{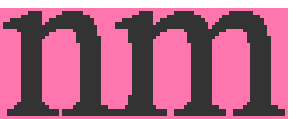}}\\
  \vspace{.1in}
  \subfigure[Close-up view of the image trapped by JTHBCT03. In order to prevent the appearance of halo artifacts, the color planes are extended underneath the gray letters by $1$ or $2$ pixels.  This results in a reddish outline around the gray areas. For easier viewing, the overall $K$ component of the image has been
  reduced. This lightens both the character region and the
  background.
  ]{\label{sf2:trask}\includegraphics[width=0.8\textwidth]{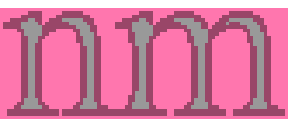}}\\
  \vspace{.1in}
  \subfigure[Close-up view of the image trapped using our proposed Algorithm 3. The center pixels in the yellow squares represent the pixels trapped by our hybrid algorithm but missed by JTHBCT03. Other than these additional trapped pixels, the overall result of Algorithm 3 is the same as that of JTHBCT03. ]{\label{sf:ours}\includegraphics[width=0.8\textwidth]{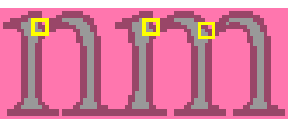}}\\
  \caption{Comparison of trapping details between JTHBCT03 and the proposed hybrid algorithm (Algorithm 3). }
  \label{fig:trapping comparison}
\end{figure}

The experimental results for the hybrid approach, along with the two
LUT-based approaches previously discussed, are tabulated in Table
\ref{tb:running time comparison}. For better comparison, we also
included the average running time obtained by combining JTHBCT03
with our prescreening rules. As can be seen from the table, the
hybrid approach is faster than JTHBCT03, with an average speed gain
of a factor greater than four. It is also significantly faster than
JTHBCT03 with prescreening. The time reduction for the hybrid
approach is especially significant when the test image contains a
lot of white area and uniform regions. For images containing
pictures or more continuous tone objects such as $vacation2.tif$,
the advantage of the hybrid approach is somewhat less. This is most
likely because we end up trapping more pixels than JTHBCT03. While
this does not seem to affect the visual quality of the trapped
image, it does slightly decrease the computational advantage. An
example comparing trapping results is shown in Fig.
\ref{fig:trapping comparison}. Notice that the difference consists
of corner pixels, which are trapped with our algorithm but not with
JTHBCT03. Since these corner pixels are located immediately next to
the character edges, it seems that they should be trapped. In any
case, the visual quality of the image is not affected by single
pixel value changes.

%% file: conclusion.tex
\section{Conclusion}
\label{conclusion}

Color trapping is an important image processing problem for color
printing that has not previously been addressed in the scholarly
research literature. We developed three different automatic color
trapping algorithms based on JTHBCT03, a previously reported color
trapping algorithm that was developed for hardware implementation.
All of our proposed algorithms are amenable to software
implementation, as they are significantly simpler computationally,
both in terms of complexity (i.e., number of ``if'' statements,
additions, and multiplications used per trapped pixel) and actual
running time. Because of their low memory requirements, two of them
are also amenable to firmware implementation on low-cost
formatter-based printers.

In all three versions, we replaced mathematical computations for
various trapping parameters by LUTs to reduce the computation time.
The amount of memory required for these LUTs is small enough for all
three methods to be deveoped in software. Our three proposed
algorithms differ in the way the edge detection step (Step Four of
JTHBCT03) is performed. In the first two versions proposed, this
step is accomplished using LUTs. The first one (Algorithm 1) is a
$3\times3$ sliding window implementation which can effectively
eliminate the gap and halo artifacts caused by a color plane
misregistration of up to one pixel in extent as well as reduce the
effect of two-pixel color plane misregistration. This algorithm runs
in less than one tenth of the time of a software implementation of
JTHBCT03. In addition to being software-friendly, the low memory
requirements of this approach make it suitable for firmware
implementation on low-cost printers. The second one (Algorithm 2),
as in JTHBCT03, uses a $5\times 5$ sliding window, and can handle
color plane mis-registrations of up to two pixels in extent. It runs
in less than one third of the time of a software implementation of
JTHBCT03. The memory requirements of this approach are too high for
firmware implementation on low-cost printers, but it could be
implemented as part of the printer driver or within a stand-alone
application running on the host computer.


The third version is a hybrid approach that combines feature
extraction and LUTs to increase the speed to more than four times
that of a software implementation of JTHBCT03 while keeping the
memory requirements very low (about 725 Kbytes). In particular, a
computationally simple prescreening step is used to identify the
majority of non-trappable pixels. The computational advantage of
this approach is more pronounced when the input image contains text
and graphic objects of uniform color regions and white areas. When a
printed image contains many objects of continuous tones levels, the
advantage over the other approaches is not as large since the
prescreening step will not eliminate many non-trappable pixels.
Moreover, because of its low memory requirement, Algorithm 3 is
amenable to firmware implementation on low-cost printers, as well as
software implementation.

\begin{table*}[!t]
\renewcommand{\arraystretch}{1.3}
\begin{center}
\caption{Complexity analysis for different color trapping
approaches}\label{tb:complexity analysis}
\begin{minipage}{\linewidth}
\centering
\renewcommand{\thefootnote}{\thempfootnote}
\begin{tabular}{|c||c|c|c|c|c|c|}
\hline \hline
  & \multirow{2}{*}{JTHBCT03} & \multicolumn{2}{c|}{LUT-based, Pixel-} &
\multicolumn{2}{c|}{LUT-based, Pixel-} & Hybrid\\[-1ex]
&  & \multicolumn{2}{c|}{independent Approach} & \multicolumn{2}{c|}{dependent Approach} & Approach\footnote{Algorithm 3} \\
\hline 

Sliding window size & $5\times5$ & $3\times3$ & $5\times5$ & $3\times3$\footnote{Algorithm 1} & $5\times5$\footnote{Algorithm 2} & $5\times5$ \\
Memory required (bytes)~\footnote{Based on images of size
$6,400\times4,900$ pixels with $5$ lines stored in the memory at any given time.} & $723.48K$ & $727.19K$ & $3.71M$ & $727.19K$ & $3.71M$ & $724.28K$ \\

No. of ``if'' statements per trapped pixel & $71$ & $11$ & $36$ & $15$ & $40$ & $16$ \\

No. of additions per trapped pixel & $234$ & $13$ & $21$ & $\le13$ & $\le21$ & $5$\\

No. of multiplications per trapped pixel & $12$ & $0$ & $0$ & $0$ & $0$ & $0$ \\

\hline

\end{tabular}
\end{minipage}
\end{center}
\end{table*}

\begin{table*}[t]
\renewcommand{\arraystretch}{1.3}
\begin{center}
\caption{Running time comparison between our two LUT-based
approaches, our hybrid approach, and JTHBC03.} \label{tb:running
time comparison}

\begin{minipage}{\linewidth}
\centering
\renewcommand{\thefootnote}{\thempfootnote}

\begin{tabular}{|c|c|c|c|c|c|c|}
\hline \hline
\multirow{3}{*}{File Name} & \multirow{2}{*}{JTHBCT03} & JTHBCT03  & \multicolumn{2}{c|}{LUT-based, Pixel-} & \multicolumn{2}{c|}{Hybrid Approach} \\[-1ex]
&  & with Pre-Screening & \multicolumn{2}{c|}{dependent Approach} & \multicolumn{2}{c|}{(Algorithm 3)}\\
\cline{2-7}
\multirow{1}{*}{($.tif$)} & Running & Running & Running & Running & Running & Edge Classification\\[-1ex]
& Time ($secs$)\footnote{The tests were performed on a Linux machine
with Intel\textregistered \textrm{ }Xeon(TM) processor and CPU speed
of $3.60\textrm{ }GHz$. Each image was tested independently
ten times and the running time was computed as the average.} & Time ($secs$)\footnotemark[\value{mpfootnote}] & Time ($secs$)\footnotemark[\value{mpfootnote}] & Time ($secs$)\footnotemark[\value{mpfootnote}] & Time ($secs$)\footnotemark[\value{mpfootnote}] & Error Rate($\%$)\footnote{This represents the difference between our trapping results and those obtained by JTHBCT03.}\\
\hline 

Window Size & $5\times5$ & $5\times5$ & $3\times3$ & $5\times5$ & $5\times5$ & ---\\
\hline 

$dino$ & $26.82$ & $10.33$ & $2.29$ & $7.02$ & $6.03$ & $0.26$ \\

$dna$  & $25.4$  & $4.02$ & $1.76$ & $3.38$ & $2.86$ & $0.08$ \\

$important$ & $26.76$ & $11.66$ & $2.62$ & $8.08$ & $6.96$ & $0.10$ \\

$textile$ & $24.33$ & $9.62$ & $2.85$ & $8.30$ & $7.25$ & $0.05$ \\

$vacation2$ & $25.86$ & $9.63$ & $2.53$ & $7.28$ & $6.08$ & $7.18$ \\

\hline

$Average$ & $25.83$ & $9.05$ & $2.41$ & $6.81$ & $5.84$ & $1.53$ \\

\hline
\end{tabular}
\end{minipage}
\end{center}
\label{table: overall time comparison}
\end{table*}

All three of our proposed trapping algorithms
 provide a flexible structure for feature selection and training
set collection, which could be easily modified for different
trapping needs.